\documentclass{elsart}

\usepackage{graphics}
\usepackage{graphicx}

\usepackage{amssymb}

\usepackage{t1enc}
\usepackage[latin1]{inputenc}

\begin{document}

\begin{frontmatter}



\title{Dialectical Multispectral Classification of Diffusion-Weighted Magnetic Resonance Images as an Alternative to Apparent Diffusion Coefficients Maps to Perform Anatomical Analysis}


\author[dsc,dee]{W. P. Santos\corauthref{cor1}},
\ead{wellington@dsc.upe.br}
\author[dee]{F. M. Assis},
\author[df]{R. E. Souza},
\author[ncsu]{P. B. Santos Filho},
\author[dsc]{F. B. Lima Neto}

\address[dsc]{Universidade de Pernambuco, Escola Politécnica de Pernambuco, Madalena, Recife, PE, 50720-001, Brazil}
\address[dee]{Universidade Federal de Campina Grande, Departamento de Engenharia Elétrica, Bodocongó, Campina Grande, PB, 58109-970, Brazil}
\address[df]{Universidade Federal de Pernambuco, Departamento de Física,\\ Cidade Universitária, Recife, PE, 50670-901, Brazil}
\address[ncsu]{North Carolina State University, Department of Physics,\\ Raleigh, North Carolina, USA}

\corauth[cor1]{Corresponding author.}

\begin{abstract}
Multispectral image analysis is a relatively promising field of research with applications in several areas, such as medical imaging and satellite monitoring. A considerable number of current methods of analysis are based on parametric statistics. Alternatively, some methods in Computational Intelligence are inspired by biology and other sciences. Here we claim that Philosophy can be also considered as a source of inspiration. This work proposes the Objective Dialectical Method (ODM): a method for classification based on the Philosophy of Praxis. ODM is instrumental in assembling evolvable mathematical tools to analyze multispectral images. In the case study described in this paper, multispectral images are composed of diffusion-weighted (DW) magnetic resonance (MR) images. The results are compared to ground-truth images produced by polynomial networks using a morphological similarity index. The classification results are used to improve the usual analysis of the apparent diffusion coefficient map. Such results proved that gray and white matter can be distinguished in DW-MR multispectral analysis and, consequently, DW-MR images can also be used to furnish anatomical information.
\end{abstract}

\end{frontmatter}

\section{Introduction}

The dialectical conception of reality is a sort of philosophical investigative method to analyze processes in nature and in human societies. Its origins are connected to different philosophies of the ancient civilizations of Greece, China and India, closely related to the thoughts of Heraclite, Plato, and the philosophies of Confucionism, Buddhism, and Zen. As a general analysis method, dialectics has experienced considerable progress due to the development of German Philosophy in the 19th century, with Hegel's dialectics and, in the 20th century, the works of Marx, Engels, and Gramsci. All those philosophers produced seminal works on the dynamics of contradictions in nature and class-based societies, giving rise to the Historical Materialism \cite{marx1980_1,engels1975,gramsci1992_1,gramsci1992_2,bobbio1990}.

The dialectical method of Historical Materialism is a tool to study systems by considering the dynamics of their contradictions, as dynamic processes with intertwined phases of \emph{evolution} and \emph{revolutionary crisis}. It has inspired us to conceive an evolvable computational intelligent method for classification that is able to solve problems commonly approached by neural networks and genetic algorithms.

Each of the most common paradigms of Computational Intelligence, namely neural networks, evolutionary computing, and culture-inspired algorithms, has its basis in a kind of theory intended to be of general application, but in fact very incomplete; e.g. the neural networks approach is based on a certain model of the brain; evolutionary computing is based on Darwin's theory; and cultural-inspired algorithms are based on the study of populations, such as those of ant colonies. However, it is important to note that it is not necessarily the case (and indeed it may be impossible) that
the theories an algorithm are based on have to be complete. For example, neural networks utilize a well-known
incomplete model of the neurons. This is a strong reason for investigating the use of Philosophy as a source of inspiration for developing computational intelligent methods and models to apply in several areas, such as pattern recognition.

Thornley and Gibb discussed the application of Dialectics to understand more clearly the paradoxical and conceptually contradictory discipline of information retrieval \cite{thornley2007}, while Rosser Jr. attempted to use some aspects of Dialectics in nonlinear dynamics, comparing some aspects of Marx and Engel's dialectical method with concepts of Catastrophe Theory, Emergent Dynamics Complexity and Chaos Theory \cite{rosser2000}. However, there are no works proposing a mathematical approach to establish the fundamentals of Dialectics as a tool for constructing computational intelligent methods.

This work presents the Objective Dialectical Method (ODM), which is an evolvable computational intelligent method, and the Objective Dialectical Classifier (ODC), an instance of ODM that operates as a non-supervised self-organized map dedicated to pattern recognition and classification. ODM is based on the dynamics of contradictions among dialectical poles. In the task of classification, each class is considered as a dialectical pole. Such poles are involved in pole struggles and affected by revolutionary crises, when some poles may disappear or be absorbed by other ones. New poles can emerge following periods of revolutionary crisis. Such a process of pole struggle and revolutionary crisis tends to a stable system, e.g. a system corresponding to the clusterization of the original data. As a case study, we use ODC to classify magnetic resonance multispectral images as an option for improving diagnosing by imaging. Notice that we are not concerned whether Dialectics is a closed model, i.e. a complete theory. We are just interested in demonstrating through an important application that Dialectics can be useful for building computational intelligent models and tools, e.g. pattern classifiers.

Alzheimer's disease is the most common cause of dementia, both in senile and presenile individuals, observing the gradual progress of the disease as the individual becomes older \cite{ewers2006,carmichael2005,hirata2005,pannacciulli2006}. The major manifestation of Alzheimer's disease is the diminution of the cognitive functions with gradual loss of memory, including psychological, neurological and behavioral symptoms indicating the decline of the daily life activities as a whole. Alzheimer's disease is characterized by the reduction of gray matter and the growth of cerebral sulci. However, the white matter is also affected, although the relation between Alzheimer's disease and white matter is still unknown \cite{friman2006,naggara2006,bozzali2002,du2005}.

Acquisition of diffusion-weighted magnetic resonance images (DW-MR images) turns possible the visualization of the dilation of the lateral ventriculi temporal corni, enhancing the augment of sulci, related to the advance of Alzheimer's disease \cite{naggara2006,haacke1999}. Therefore, volumetrical measuring of cerebral structures is very important for diagnosis and evaluation of the progress of diseases like Alzheimer's \cite{ewers2006,carmichael2005,hirata2005,pannacciulli2006}, especially the measuring of the volumes occupied by sulci and lateral ventriculi, turning possible the addition of quantitative information to the qualitative information expressed by the DW-MR images \cite{hayasaka2006}.

Usually, the evaluation of the progress of Alzheimer's disease using image analysis of DW-MR images is performed after  acquiring at least three images of each slice of interest, generated using the sequence spin-echo Stejskal-Tanner with different diffusion exponents, where one of the exponents is 0 s/mm$^2$, that is, a $T_2$-weighted spin-echo image \cite{haacke1999}. Then, a fourth image is calculated: the Apparent Diffusion Coefficient Map, or ADC map, where each pixel is associated to the corresponding apparent diffusion coefficient of the associated voxel: the brighter the pixels, the greater the corresponding apparent diffusion coefficients \cite{haacke1999}.

According to several works, it relatively common to assume that it is not possible to make distinction between gray and white matter in DW-MR images \cite{santos2007d,santos2008r,santos2007a,santos2007b,santos2007c,santos2006a,santos2006b}. This work proves that is possible to distinguish gray and white matter by the use of non-supervised adaptative classifiers, specially using the dialectical objective classifier proposed in this work, to classify the synthetic multispectral classifier composed by the several acquired DW-MR images, making possible to visualize the reduction of gray matter in frontal lobule, where recent memory is located. Its fundamental to measure such an area of gray matter to evaluate the progress of diseases as Alzheimer's.

This work proposes a relatively new approach to evaluate diseases using DW-MR images: once the ADC map usually presents pixels with considerable intensities in regions not occupied by the head of patient, a degree of uncertainty can also be considered in the pixels inside the sample. Furthermore, the ADC map is very sensitive to noisy images \cite{haacke1999,santos2007d}. Therefore, in this case study, images are used to compose a multispectral image, where each DW-MR image is considered as a spectral band in a synthetic multispectral image. This multispectral image is classified using the Objective Dialectical Classifier, a new classification method based on Dialectics as defined in the Philosophy of Praxis.

This paper is organized as follows: in section \ref{sec_methods} the data obtained from the case study in Alzheimer's and the multispectral classification methods are presented. Specifically, the image database used in this study is presented in subsection \ref{subsec_images}. The objective dialectical method is described in details in subsection \ref{subsec_dialectics}. Classification methods based on neural networks, fuzzy c-means maps, and objective dialectical classifiers are shown in subsections \ref{subsec_multspec}, \ref{subsec_fuzzy}, and \ref{subsec_dialetica}, respectively. Computational tools employed in this work are shown in subsection \ref{subsec_comptools}. Aspects on granulometry are commented in subsection \ref{subsec_granu}. The morphological similarity index used to evaluate part of our results is described in subsection \ref{subsec_morphindex}. Classification results are presented in section \ref{sec_results}. Finally, discussions and conclusions are made in section \ref{sec_discussion}.

\section{Materials and Methods} \label{sec_methods}

\subsection{DW-MR Images and ADC Maps} \label{subsec_images}

The diffusion-weighted magnetic resonance (MR) images were acquired from the clinical images database of the Laboratory of MR Images, at the Physics Department of Universidade Federal de Pernambuco, Recife, Brazil. The image database is composed by real clinical images acquired from a clinical MR tomographer of 1.5 T. The MR images used in this work corresponds to an unique volunteer with Alzheimer's. As a case study there were used 80 DW-MR images, corresponding to 4 volumes of 20 axial slices, where one of these volumes is composed by ADC maps. All of these images correspond to a 70-year-old male patient with Alzheimer's disease, with the following diffusion exponents: 0 s/mm$^2$, 500 s/mm$^2$ and 1000 s/mm$^2$.
\begin{figure}
    \centering
        \includegraphics[width=0.45\textwidth]{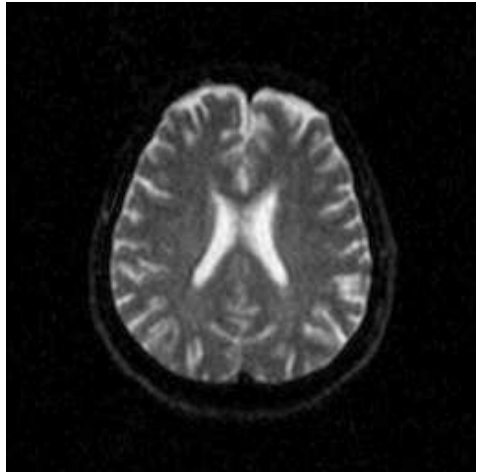}
    \caption{Axial DW-MR image of 13th slice with diffusion exponent of 0 s/mm$^2$}
    \label{fig:t1_13}
\end{figure}
\begin{figure}
    \centering
        \includegraphics[width=0.45\textwidth]{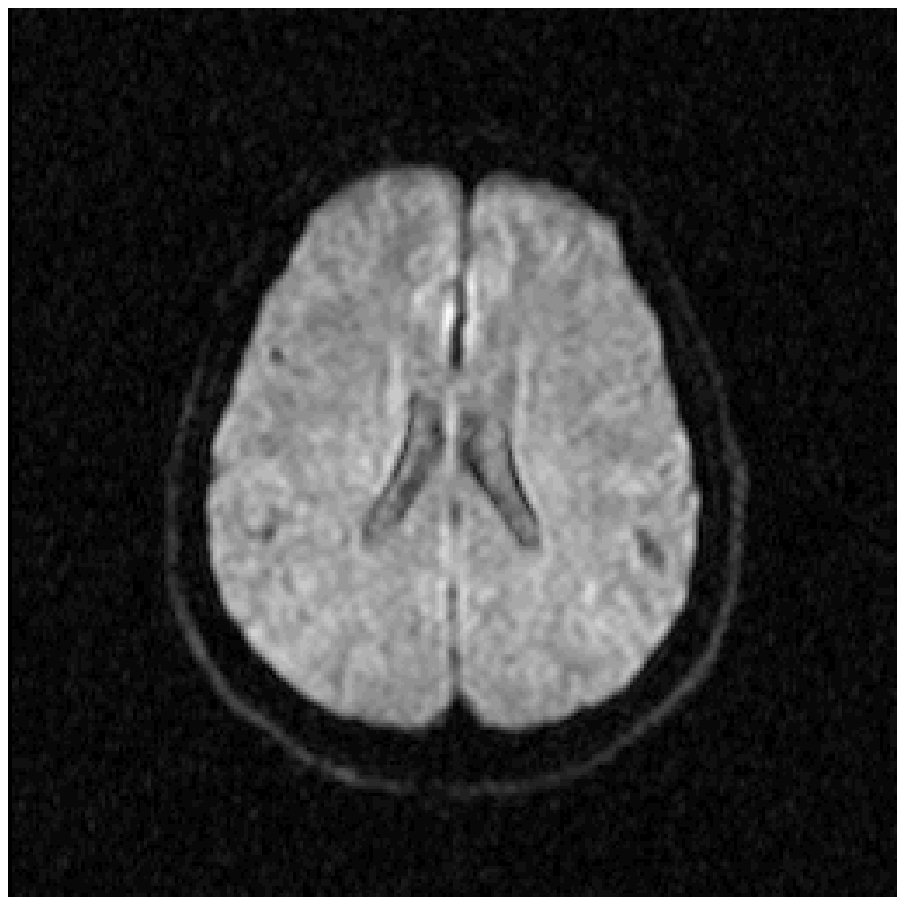}
    \caption{Axial DW-MR image of 13th slice with diffusion exponent of 500 s/mm$^2$}
    \label{fig:t2_13}
\end{figure}
\begin{figure}
    \centering
        \includegraphics[width=0.45\textwidth]{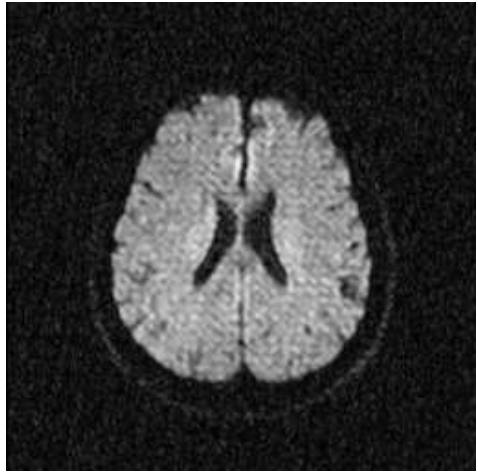}
    \caption{Axial DW-MR image of 13th slice with diffusion exponent of 1000 s/mm$^2$}
    \label{fig:t3_13}
\end{figure}

To perform the training we chose 13th slice of each volume sample (figures \ref{fig:t1_13}, \ref{fig:t2_13} and \ref{fig:t3_13}), once this slice shows the temporal corni of the lateral ventriculi. The exhibition of such structures facilitates the analysis of the specialist and helps him to find a correlation between data generated by our computational tool and \emph{a priori} specialist knowledge. Furthermore, slice 13 presents a considerable amount of artifacts out of the cranial region.
\begin{figure}
    \centering
        \includegraphics[width=0.45\textwidth]{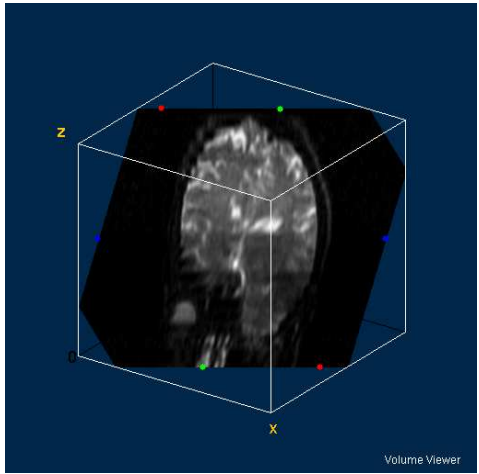}
    \caption{DW-MR volume with diffusion exponent of 0 s/mm$^2$}
    \label{fig:VistaVolumet1}
\end{figure}
\begin{figure}
    \centering
        \includegraphics[width=0.45\textwidth]{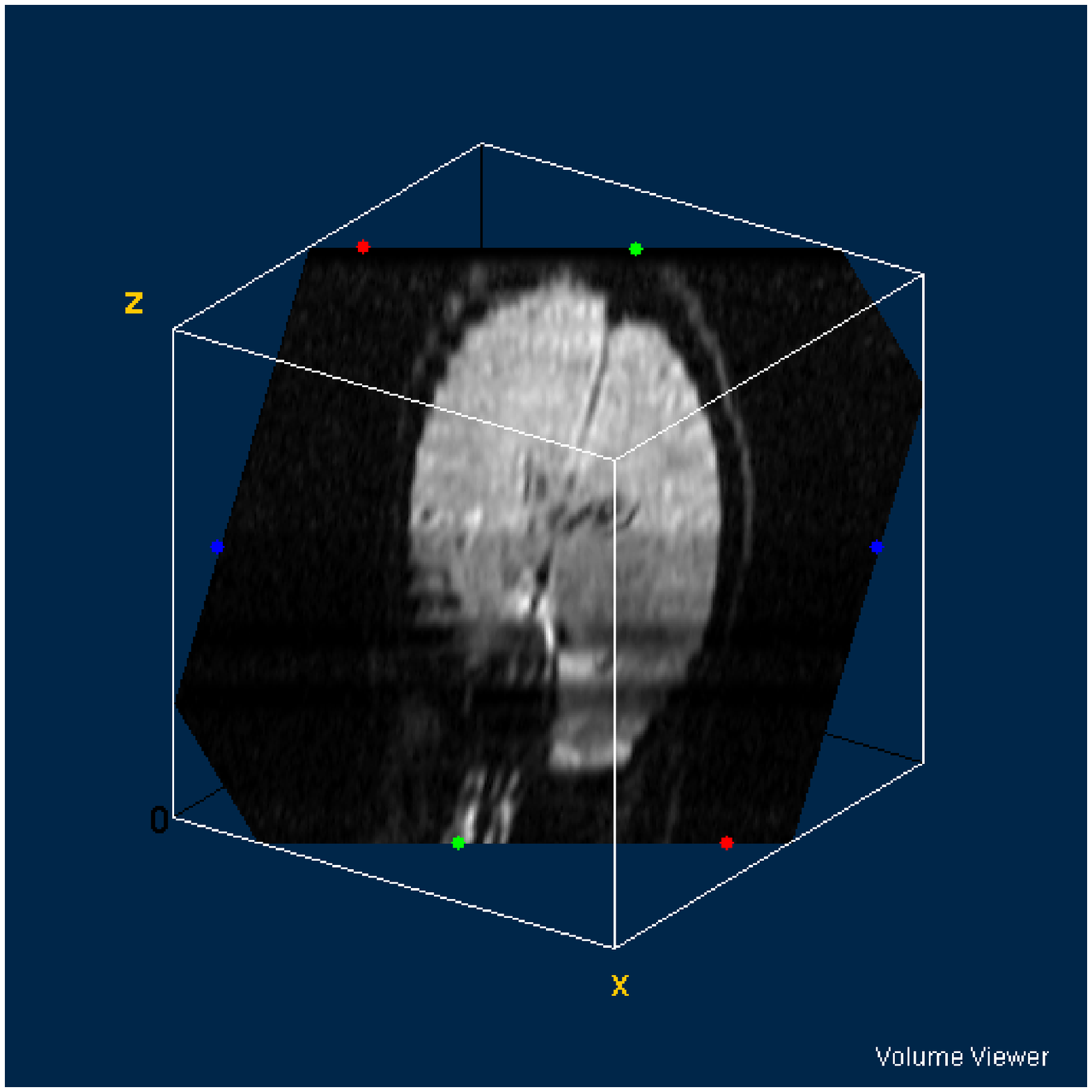}
    \caption{DW-MR volume with diffusion exponent of 500 s/mm$^2$}
    \label{fig:VistaVolumet2}
\end{figure}
\begin{figure}
    \centering
        \includegraphics[width=0.45\textwidth]{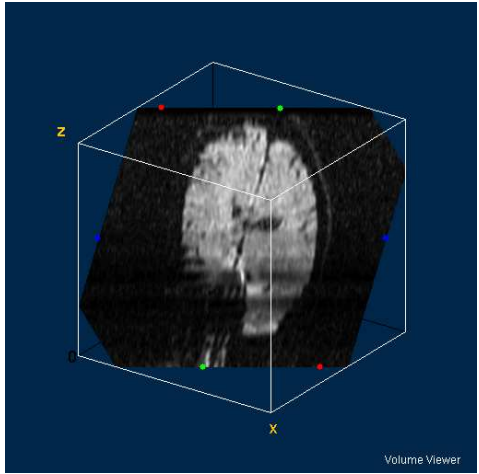}
    \caption{DW-MR volume with diffusion exponent of 1000 s/mm$^2$}
    \label{fig:VistaVolumet3}
\end{figure}

The images can be considered as mathematical functions, where their domain is a region of the plain of the integers, called \emph{grid}, and their counterdomain is the set of the possible values occupied by the pixels corresponding to each position of the grid.

Let $f_i:S\rightarrow W$ be the set of the diffusion-weighted MR images, where $1\leq i\leq 3$, $S\subseteq \textbf{Z}^2$ is the grid of the image $f_i$, where $W\subseteq \textbf{R}$ is its counterdomain. The synthetic multispectral image $f:S\rightarrow W^3$ corresponding to a determined slice of the volume composed by the DW-MR volumes presented in figures \ref{fig:VistaVolumet1}, \ref{fig:VistaVolumet2} and \ref{fig:VistaVolumet1} is given by:
\begin{equation}
  f(\textbf{u})=(f_1(\textbf{u}),f_2(\textbf{u}),f_3(\textbf{u}))^T,
\end{equation}
where $\textbf{u}\in S$ is the position of the pixel in the image $f$, and $f_1$, $f_2$ and $f_3$ are the diffusion-weighted MR images with diffusion exponents of $b_1=0$ s/mm$^2$, $b_2=500$ s/mm$^2$, and $b_3=1000$ s/mm$^2$, respectively.

The analysis of diffusion-weighted MR images is often performed using the resulting ADC map. Considering that each pixel $f_i(\textbf{u})$ is approximately proportional to the signal of the corresponding voxel as follows \cite{wu2001,elshafiey2002,basser2003,wang2004,chen2004,guo2006,maraga2006}:
\begin{equation}
f_i(\textbf{u})=K\rho(\textbf{u})e^{-T_E/T_2(\textbf{u})}e^{-b_i D_i(\textbf{u})},
\end{equation}
where $D_i(\textbf{u})$ is the diffusion coefficient associated to the voxel mapped in the pixel in the position $\textbf{u}$, $\rho(\textbf{u})$ is the nuclear spin density in the voxel, $K$ is a constant of proportionality, $T_2(\textbf{u})$ is the transversal relaxation in the voxel, $T_E$ is the echo time and $b_i$ is the diffusion exponent, given by \cite{haacke1999,liang2000,gullberg1999}:
\begin{equation}
b_i=\gamma^2 G_i^2 T_E^3/3,
\end{equation}
where $\gamma$ is the gyromagnetic ratio and $G_i$ is the gradient applied during the experiment $i$; the ADC map $f_{\texttt{ADC}}:S\rightarrow W$ is calculated as follows \cite{basser2002}:
\begin{equation} \label{eq:fadc}
	f_{\texttt{ADC}}(\textbf{u})=\frac{C}{b_2}\ln\left( \frac{f_1(\textbf{u})}{f_2(\textbf{u})} \right) + \frac{C}{b_3}\ln\left( \frac{f_1(\textbf{u})}{f_3(\textbf{u})} \right),
\end{equation}
where $C$ is a constant of proportionality. Thus, the ADC map is given by:
\begin{equation}
  f_{\texttt{ADC}}(\textbf{u})=CD(\textbf{u}).
\end{equation}

Considering $n$ experiments, we could generalize equation \ref{eq:fadc} as follows:
\begin{equation} \label{eq:fadcg}
    f_{\texttt{ADC}}(\textbf{u})=\sum_{i=2}^n \frac{C}{b_i} \ln\left( \frac{f_1(\textbf{u})}{f_i(\textbf{u})} \right).
\end{equation}

Therefore, the ADC map is given by:
\begin{equation}
  f_{\texttt{ADC}}(\textbf{u})=C\bar{D}(\textbf{u}),
\end{equation}
where $\bar{D}(\textbf{u})$ is the sample average of the measurements of the diffusion coefficient $D(\textbf{u})$ \cite{basser2003,kang2005,fillard2006}.

\begin{figure}
    \centering
        \includegraphics[width=0.45\textwidth]{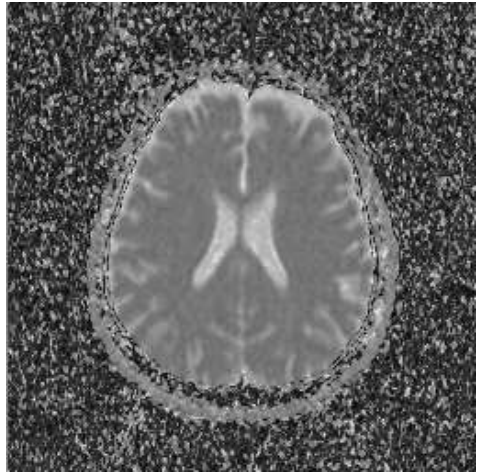}
    \caption{ADC map of 13th slice calculated from the three diffusion images presented in figures \ref{fig:t1_13}, \ref{fig:t2_13} and \ref{fig:t3_13}}
    \label{fig:ADC_13}
\end{figure}

\begin{figure}
    \centering
        \includegraphics[width=0.45\textwidth]{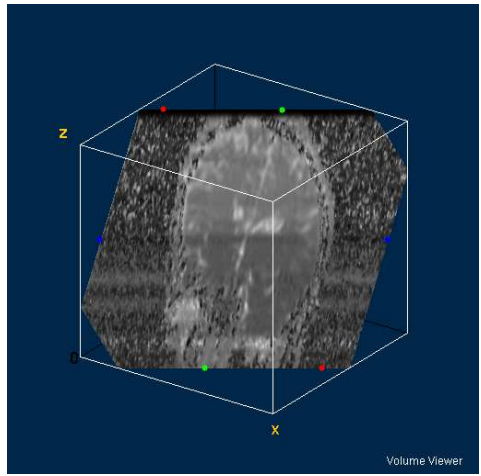}
    \caption{ADC volume calculated from the volumes of figures \ref{fig:VistaVolumet1}, \ref{fig:VistaVolumet2} and \ref{fig:VistaVolumet3}}
    \label{fig:VistaVolumeADC}
\end{figure}

Therefore, the pixels of the ADC map are proportional to the diffusion coefficients in the corresponding voxels. However, as the images are acquired at different moments, there must be considered the occurrence of noise in all the experiments. Furthermore, the presence of noise is amplified by the use of the logarithm. In figures \ref{fig:ADC_13} and \ref{fig:VistaVolumeADC} it is possible to see several artifacts generated by the presence of noise. In regions where signal-to-noise ratio is poor (e.g. $s/n \approx 1$), the ADC map produces artifacts as a consequence of the calculation of logarithms (notice equations \ref{eq:fadc} and \ref{eq:fadcg}).

Such factors leave us to the following conclusion: the pixels of the ADC map not necessarily correspond to the diffusion coefficients: several pixels indicate high diffusion rates in voxels where the sample are not present or in very solid areas like bone in the cranial box, as can be seen in figures \ref{fig:ADC_13} and \ref{fig:VistaVolumeADC}. This is the reason why such map indicates \emph{apparent} diffusion coefficients, and not \emph{real} diffusion coefficients.

In this work we propose an alternative to the analysis of ADC maps: the multispectral analysis of volumes composed by images $f:S\rightarrow W^3$ using classification methods based on neural networks and objective dialectical classifiers.

\subsection{The Objective Dialectical Method} \label{subsec_dialectics}

The Objective Dialectical Method (ODM) is an evolvable computational intelligent method designed to model dynamic systems and to perform tasks of classification, pattern recognition, intelligent search and optimization \cite{santos2008a,santos2008b}.

General principles of ODM are as the following algorithm:
\begin{enumerate}
	\item System inputs must be represented as a \emph{vector of conditions} representing the main features of the problem;
	\item The user has to provide the initial parameters for: \emph{poles} or \emph{classes} that compose the system, \emph{number of historical phases}, and \emph{duration of each historical phase}. The number of historical phases and their duration can also be randomly defined;
	\item Each dialectical pole is associated with: (i) a \emph{vector of weights}, with the same size as the vector of conditions; (ii) an \emph{anticontradiction function}, and (iii) a \emph{measure of force}. Such a vector of weights can be randomly defined or chosen from the set of vectors of conditions;
	\item The historical phases consist of two stages:
	\begin{enumerate}
		\item \textbf{Evolution:} Where conditions are presented to the inputs of the system and the integrating poles compete with each other, in a process called \emph{pole struggle}. Then, an anticontradiction function associated with each pole is evaluated and, given a vector of conditions, the winner pole, which is the pole with greatest anticontradiction value, has its parameters (weights and measure of force) incremented. This process continues until the end of the historical phase is reached;
		\item \textbf{Revolutionary crisis:} This starts at the end of the historical phase. At this point the following steps are performed:
		\begin{enumerate}
			\item The measures of force are compared, i.e. all the poles with a measure of force less than a minimum force are marked;
			\item The con\-tra\-dic\-tions among the integrating po\-les are also evaluated. If a contradiction between two poles is less than a given minimum contradiction, one of the two poles is selected or mar\-ked as such. Here the minimum contradiction plays the role of a threshold;
			\item From the evaluated contradictions computed in the previous items, the overall maximum contradiction is calculated. This is \emph{the main contradiction} of the system. From the pair of poles involved in the main contradiction, a new pole is generated, i.e. a synthesis of previous pair of poles, whose vector of weights is calculated from the vector of weights of the pair. It is also possible to choose more than one main contradiction and, from its involved pairs of poles, generate other new poles;
			\item All the marked poles are eliminated and a new set of integrating poles is generated;
			\item The vector of weights of all poles of the new set of poles is randomly modified, representing the impact of the revolutionary crisis on both the survivors and the new dialectical poles.
		\end{enumerate}
	\end{enumerate}
\end{enumerate}

Objective Dialectical Classifiers are based on the Objective Dialectical Me\-thod. They are an adaptation of ODM to tasks of classification. This means that the feature vectors are mounted and considered as vectors of conditions. Specifically, once they are applied to the inputs of the dialectical system, their coordinates will affect the dynamics of the contradictions among the integrating dialectical poles. Hence, the integrating poles model the recognized classes at the task of non-supervised classification.

Therefore, an objective dialectical classifier is in fact an adaptable and evolvable non-supervised classifier where, instead of supposing a predetermined number of classes, we can set an initial number of classes (dialectical poles) and, as the historical phases happen (as a result of pole struggles and revolutionary crises), some classes are eliminated, others are absorbed, and a few others are generated. At the end of the training process, the system presents a number of statistically significant classes present in the training set and, therefore, a feasible classifier associated to the final state of the dialectical system.

To accelerate the convergence of the dialectical classifier, we have removed the operator of pole generation, present at the revolutionary crises. However, it could be beneficial to the classification method, once such operator is a kind of diversity generator operator. The solution found can then be compared to other sort of evolvable classifiers.

The following algorithm is a possible implementation of the training process of the objective dialectical classifier:
\begin{enumerate}
	\item Set the following initial parameters:
		\begin{enumerate}
			\item Number of historical phases, $n_P$;
			\item Length of each historical phase, $n_H$;
			\item Desired final number of poles, $n_{C,f}$;
			\item Step of each historical phase, $0<\eta(0)<1$;
	    \item Maximum crisis, $0\leq \chi_{\max}\leq 1$;
			\item Initial number of poles $\#\Omega(0)=n_C(0)$, defining the initial set of poles:
	    $$
	    \Omega(0)=\{ C_1(0), C_2(0), \dots, C_{n_C(0)}(0) \}.
	    $$
	  \end{enumerate}
	\item Set the following thresholds:
		\begin{enumerate}
			\item Minimum force, $0\leq f_{\min}\leq 1$;
			\item Minimum contradiction, $0\leq \delta_{\min}\leq 1$;
    \end{enumerate}	
	\item Initialize the weights $w_{i,j}(0)$, where $1\leq i\leq n_C(0)$ and $1\leq j\leq n$.
	\item Let $\#\Omega(t)$ be the cardinality of $\Omega(t)$, repeat until $n_P$ iterations or $\#\Omega(t)=n_{C,f}$: 
	\begin{enumerate}
	  \item Repeat until $n_H$ iterations:
	  \begin{enumerate}
	    \item Initialize the measures of force $f_i=0$, for $1\leq i\leq n_{C}(t)$.
	    \item For all vectors of conditions
	    $$
	    \mathbf{x}=(x_1, x_2, \dots, x_n)^T
	    $$
	    of the input set $\Psi=\{\mathbf{x}^{(l)}\}_{l=1}^L$, repeat:
	    \begin{enumerate}
	      \item Compute the values of the anticontradiction functions:
	      $$ g_i(\mathbf{x})=e^{-||\mathbf{x}-\mathbf{w}_i||}, $$
	      where $1\leq i\leq n_{C}(t)$.
	      \item Calculate $g_{\max}$:
        $$ g_{\max}=\max\{g_1(\mathbf{x}), g_2(\mathbf{x}), \dots, g_{n_C(t)}(\mathbf{x})\}. $$
        \item Calculate the index $k(t)$ of the winner class:
        $$ g_i=g_{\max}\Rightarrow k(t)=i. $$	  
        \item Adjust the weights of the winner pole:
        $$
        w_{i,j}(t+1)=\left\{
          \begin{array} {ll}
          {w_{i,j}'(t),} & {i=k(t)}\\
          {w_{i,j}(t),} & {i\neq k(t)}
          \end{array}
        \right.,
        $$
        where
        $$
        w_{i,j}'(t)=w_{i,j}(t)+\eta(t)(x_j(t)-w_{i,j}(t)).
        $$
        \item Update the measure of force of the integrating poles:
        $$
        f_{i}(t+1)=\left\{
          \begin{array} {ll}
          {f_{i}(t)+1,} & {i=k(t)}\\
          {f_{i}(t),} & {i\neq k(t)}
          \end{array}
        \right..
        $$    
	    \end{enumerate}
	    \item Quantitative changing: $\Omega(t+1)=\Omega(t)$.
	  \end{enumerate}
  \item Calculate the normalized measures of force:
  $$
  \bar{f}_i(t)=\frac{f_i(t)}{\max\{f_j(t)\}_{j=1}^{n_C(t)}},
  $$
  for $1\leq i\leq n_C(t)$.
  \item Compute the contradictions:
  $$ \delta_{i,j}=1-g_i(\mathbf{w}_j), $$
  where $2\leq j\leq n_C(t)$, $1\leq i<j$, and find the maximum contradiction
  $$ \delta_{\max}=\max\{\delta_{i,j}, i\neq j\}, $$
  for $j = 2, 3, \dots, n_C(t)$ and $i = 1, 2, \dots, j-1$.
  \item Qualitative changing: compute the new set of poles, $\Omega(t+1)$:
  $$ \bar{f}_i(t)>f_{\min}\Rightarrow C_i(t)\in \Omega(t+1), $$
  where $1\leq i\leq n_C(t)$ and
  $$ \delta_{i,j}\geq\delta_{\min}\Rightarrow C_i(t), C_j(t)\in \Omega(t+1), $$
  $$ \delta_{i,j}<\delta_{\min}\Rightarrow C_i(t)\in \Omega(t+1), $$
  $$ \delta_{i,j}=\delta_{\max}\Rightarrow C_q\in \Omega(t+1), $$
  where $2\leq j\leq n_C(t)$, $1\leq i<j$, $q=n_C(t)+1$, and
  $$
  w_{q,k}(t+1)=\left\{
  \begin{array}{ll}
    {w_{i,k}(t+1),} & {k~\textnormal{mod}~2=1}\\
    {w_{j,k}(t+1),} & {k~\textnormal{mod}~2=0}
  \end{array} \right.,
  $$
  for $k=1,2,\dots,n$.
  \item Add the crisis effect to the weights of the new integrating poles of the dialectical system:
  $$ w_{i,j}(t+2)=w_{i,j}(t+1)+\chi_{\max} r(t), $$
  for $1\leq i\leq n_C(t+1)$, $1\leq j\leq n$ and $\Omega(t+2)=\Omega(t+1)$; $r(t)$ is a random Gaussian variable with distribution $r(t)\sim G(0,1)$.
  \end{enumerate}
\end{enumerate}

Once the training process is complete, objective dialectical classifier behavior occurs in the same way as any non-supervised classification method. This is clear if we analyze the training process when $n_P=n_H=1$. This transforms the ODC into a k-means method, for instance.

The classification is performed in the following way: given a set of input conditions
$$ \mathbf{x}=(x_1, x_2, \dots, x_n)^T, $$
if the dialectical system reaches stabilization when $\Omega=\{C_1, C_2, \dots, C_{n_C}\}$, then we calculate:
$$ g_{\max}=\max\{g_1(\mathbf{x}), g_2(\mathbf{x}), \dots, g_{n_C}(\mathbf{x})\}. $$
Therefore, the classification rule is:
$$ g_k(\mathbf{x})=g_{\max}\Rightarrow \mathbf{x}\in C_k, $$
where $1\leq k\leq n_C$.

\subsection{Multispectral Analysis using Neural Networks} \label{subsec_multspec}

Let the universe of classes of interest be defined as $\Omega=\{C_1,C_2,C_3\}$, $C_1$ represents the cerebrospinal fluid; $C_2$, the white and the gray matter, once they cannot be distinguished using diffusion images, because their diffusion coefficients are very close; $C_3$ corresponds to the image background.

For the multispectral analysis using neural nets, the inputs are associated to the vector $\textbf{x}=(x_1,x_2,x_3)^T$, where $x_i=f_i(\textbf{u})$, for $1\leq i\leq 3$. The net outputs represent the classes of interest and are associated to the vector $\textbf{y}=(y_1,y_2,y_3)^T$, where each output corresponds to the class with the same index. The decision criterion employed in such analysis is the Bayes criterion: the output with greater value indicates the more probable class \cite{duda2001,duda1972,sklansky1981}. The training set and the test set were built using specialist knowledge at the selection of the regions of interest \cite{haykin1999}.

The volume of multispectral images was classified using the following methods:
\begin{enumerate}
  \item \emph{Multilayer Perceptron} (MLP): Initial learning rate $\eta_0=0.2$, training error $\epsilon=0.05$, maximum number of 1000 iterations, 3 inputs, 3 outputs, 2 layers, 60 neurons in layer 1 \cite{haykin1999};
  \item \emph{Radial Basis Function Network} (RBF): 3 inputs, 2 layers; layer 1: k-means map with 18 neurons in layer 1, initial learning rate $\eta_0=0.1$, maximum of 200 iterations; layer 2:  3 outputs, maximum of 200 iterations, initial learning rate $\eta_0=0.1$ \cite{haykin1999}.
  \item \emph{Kohonen LVQ Classifier} (LVQ): 3 inputs, 3 outputs, maximum of 200 iterations, initial learning rate $\eta_0=0.1$ \cite{haykin1999}.
  \item \emph{Kohonen Self-Organized Map} (KO): 3 inputs, 3 outputs, maximum of 200 iterations, initial learning rate $\eta_0=0.1$ \cite{haykin1999}.
  \item \emph{Polynomial Network} (PO): Two-layer network: the first layer is a multiplicative net that generates the terms of the 2-degree polynomial from the 3 inputs, and the second layer consists on an one-layer perceptron with learning rate $\eta=0.1$ and training error $\epsilon=0.05$, maximum of 200 training iterations, responsible for the calculation of the coefficients of the polynomial that models the discriminant polynomial function of each class \cite{duda2001,sklansky1981}.
\end{enumerate}

The multilayer perceptron was chosen to evaluate the performance of the multispectral classification based on classical neural networks. The numbers of inputs and outputs correspond to the numbers of bands and classes, respectively. The training error was defined by considering the maximum estimated noise in diffusion-weighted images. The number of neurons in layer 1 and the learning rate were determined empirically.

The radial basis function network was chosen to aid to evaluate of the performance of multispectral classification based on a local problem-oriented strategy. The number of inputs and outputs correspond to the number of bands and classes of interest, respectively. The initial learning rate was determined empirically.

The Kohonen self-organized map was chosen to aid at the evaluation of the performance of a non-supervised multispectral classifier based on clustering. However, the Kohonen LVQ classifier was chosen as a supervised alternative to the multispectral classification performed by the use of multilayer perceptrons and radial basis function networks.

\subsection{Multispectral Analysis using Fuzzy C-Means Maps} \label{subsec_fuzzy}

The \emph{fuzzy c-means map} (CM) was also used to perform the classification of the synthetic multispectral images generated from diffusion images, similarly to neural networks. There was used a fuzzy c-means map with 3 inputs and 3 outputs, trained during a maximum of 200 iterations with an initial learning rate $\eta_0=0.1$.

\subsection{Multispectral Analysis using the Dialectical Classifier} \label{subsec_dialetica}

The \emph{objective dialectical classifier} (ODC) was trained using an initial system with 10 integrating classes submitted to 3 input conditions and studied for 5 historical 100-iteration phases, with historical step $\eta_0=0.1$. At the stages of revolutionary crises we considered minimum force of 1\%, minimum contradiction of 25\%, and maximum crisis of 25\%. The proposed algorithm runs until the final number of classes, that is, 4 classes, is reached. The input conditions are pixel values in each band. In this case we have 3 bands.

To accelerate the convergence of the dialectical classification, we did not use the operator of generation of new classes in revolutionary crisis, used during the training process. The anticontradiction functions were defined as follows:
\begin{equation}
  g_i(\mathbf{x})=\left(\sum_{k=1}^{n_C} \frac{||\mathbf{x}-\mathbf{w}_i||^2} {||\mathbf{x}-\mathbf{w}_k||^2} \right)^{-1},
\end{equation}
where $1\leq i\leq n_C$. The adjustment of weights in evolution stage, for each historical phase, was performed as in the following expression:
\begin{equation}
  w_{i,j}(t+1)=\left\{
  \begin{array} {ll}
      {w_{i,j}(t)+\eta(t) g_i^2(\mathbf{x}(t))(x_j(t)-w_{i,j}(t)),} & {i=k}\\
      {w_{i,j}(t),} & {i\neq k}
    \end{array}
  \right.,
\end{equation}
where $\eta(t)$ is the historical step and can be reduced as the process of training evolves, similarly to the way the learning rate is updated in the process of training Kohonen self-organized networks \cite{haykin1999}.

\subsection{Computational Tools} \label{subsec_comptools}

To implement the proposed and used methods and rebuild volumes, we developed \emph{AnImed}, a computational tool build using Object Pascal and Delphi 5 IDE. The volumes of interest were visualized using \emph{ImageJ}, a Java-based tool developed by NIH (\emph{National Institute of Health}, USA) and plugin \emph{VolumeViewer}.

\subsection{Granulometry} \label{subsec_granu}

Amongst the tools of the Mathematical Morphology for the description of size and form in analysis of images, the \emph{granulometries} constitute some of the most powerful tools.

The basic idea of granulometry is to perform boltings followed of quantitative measurements of the residues, such as measurement of area or perimeter. Consequently, the original image is bolted using a certain bolter and afterwards the grains are measured; after that, a bolter with lesser punctures is applied and the residues are measured again and so on. The size of the punctures of the bolter is determined by the dimension of the structure element used in the granulometric transformation.

The family of transforms $\{\psi_j\}$, parametrized using parameter $j\geq 0$, where $\psi_0(f)(u)=f(u),~\forall u\in S,~f:S\rightarrow [0,1]$, is called a granulometry if it is a \emph{shape criterion}, as in following definition \cite{ulisses1994}:
\begin{eqnarray}\label{mmeq32}
\psi_j(f)(u)\leq f(u),\\
f(u)\leq g(u)\Rightarrow\psi_j(f)(u)\leq\psi_j(g)(u),\\
\psi_j[\psi_k(f)](u)=\psi_k[\psi_j(f)](u)=\psi_{\max(j,k)}(f)(u),~~\forall
j,k\geq 0,
\end{eqnarray}
$\forall u\in S$, where $g:S\rightarrow [0,1]$.

The last condition express the following intuitive idea: two consecutive boltings are equivalent to an unique bolting using lesser punctures ($\max(j,k)$), once index $j$ of transform $\psi_j$ express the degree of bolting: the bigger $j$, the stronger the bolting and the lesser the residual grains.

It is possible to demonstrate that the set of morphological transforms able to classified as a \emph{shape criterion} is the \emph{set of $j$-openings} by structure element \emph{digital disc} \cite{ulisses1994}. The basic digital discs are square $3\times 3$ and cross $3\times 3$ \cite{ulisses1994}. Thus $\{\gamma_g^j\}$ is a \emph{granulometry}, where $j\geq 0$ and $g$ is a digital disc.

Therefore, a granulometry is now expressed by $\{\psi_j=\gamma_g^j\},~r\geq 0$, where $g$ is a digital disc \cite{ulisses1994}.

Comparatively, we can define an \emph{anti-granulometry} as the set of $j$-closings $\{\phi_g^j\}$, where $j\geq 0$ and $g$ is a digital disc \cite{ulisses1994}.

Let us consider image $f:S\rightarrow [0,1]$. Function $V:\mathbb{Z}_+\rightarrow \mathbb{R}$ is defined as follows \cite{ulisses1994}:
\begin{equation} \label{mmeq33}
V(k)=\sum_{u\in S}\gamma_g^k(f)(u),
\end{equation}
where $g$ can be cross or square $3\times 3$. Hence, function $\Xi:\mathbb{Z}_+\rightarrow [0,1]$ is defined as follows:
\begin{equation} \label{mmeq34}
\Xi[k]=1-\frac{V(k)}{V(0)},~~k\geq 0.
\end{equation}

Considering image $f$ as a \emph{random set} and emphasizing that $\Xi$ is a function monotonic, increasing and limited to interval $[0,1]$, once $V(k+1)<V(k),~\forall k\geq 0$, we can affirm that $\Xi$ is the \emph{discrete accumulated density function} associated to image $f$ \cite{ulisses1994}. Thus we can define the \emph{discrete density function} $\xi:\mathbb{Z}_+\rightarrow \mathbb{R}_+$ using the following difference:
\begin{equation}\label{mmeq35}
\xi[k]=\Xi[k+1]-\Xi[k],~~k\geq 0.
\end{equation}

The density function defined in equation \ref{mmeq35} is called \emph{pattern spectrum} \cite{ulisses1994}. The pattern spectrum, or \emph{morphological spectrum}, is a sort of \emph{size and shape histogram}. It is \emph{unique} for a specific image $f$ and a determined structure element $g$, since $f$ is a \emph{binary} image \cite{ulisses1994}.

\begin{figure}
\begin{center}
  \includegraphics[width=0.45\textwidth]{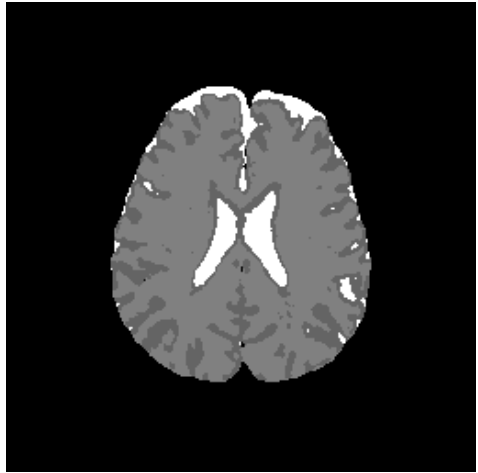}\\
  \caption{Axial image of human brain}\label{fig_mmimg_orig}
\end{center}
\end{figure}
\begin{figure}
\begin{center}
  \includegraphics[width=0.45\textwidth]{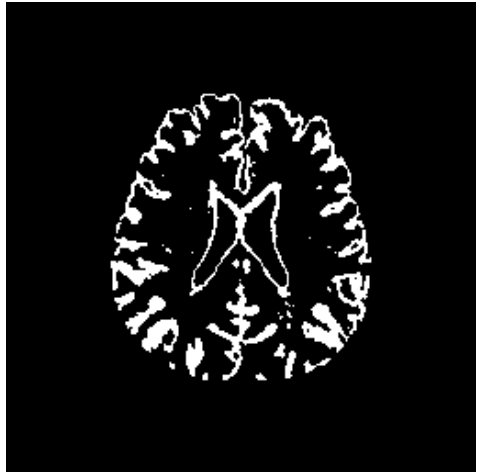}\\
  \caption{Binarized version of image of figure \ref{fig_mmimg_orig} using a threshold of 0.4}
  \label{fig_mmimg_bin}
\end{center}
\end{figure}
\begin{figure}
\begin{center}
  \includegraphics[width=0.9\textwidth]{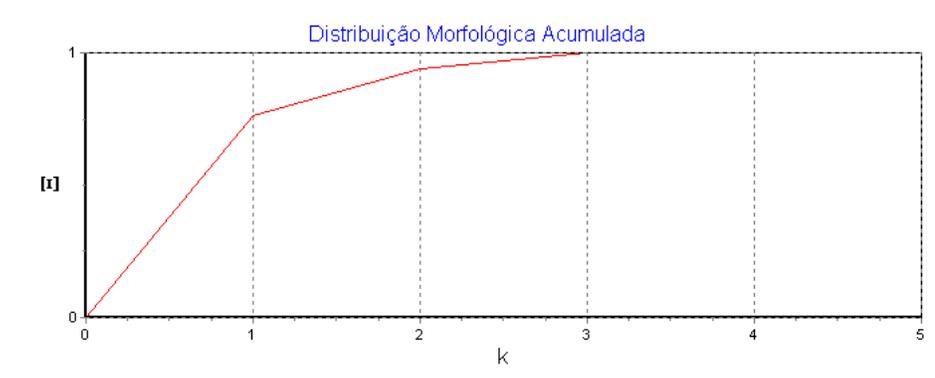}\\
  \caption{Discrete accumulated distribution function $\Xi$ of image of figure \ref{fig_mmimg_bin}}\label{fig_GraficoDistribuicao}
\end{center}
\end{figure}
\begin{figure}
\begin{center}
  \includegraphics[width=0.9\textwidth]{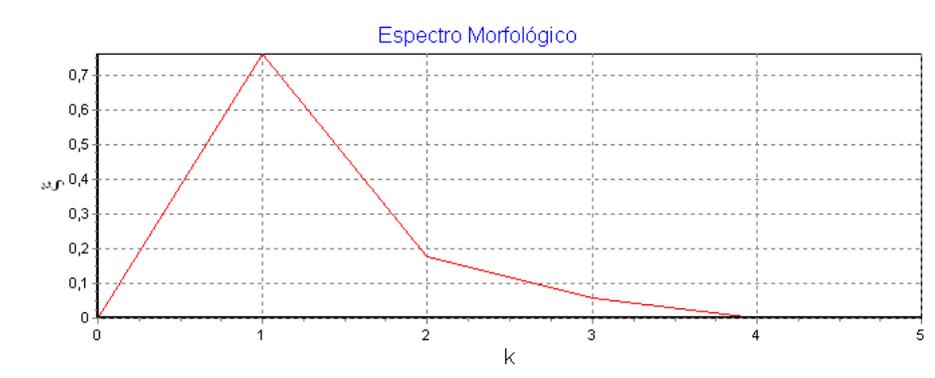}\\
  \caption{Pattern spectrum $\xi$ of image of figure \ref{fig_mmimg_bin}, obtained from the discrete derivative of accumulated distribution function plotted in figure \ref{fig_GraficoDistribuicao}}\label{fig_GraficoEspectro}
\end{center}
\end{figure}
Figures \ref{fig_GraficoDistribuicao} and \ref{fig_GraficoEspectro} show graphics of the \emph{discrete accumulated distribution function} $\Xi$ and of the \emph{pattern spectrum} $\xi$, calculated from image of figure \ref{fig_mmimg_bin}, respectively.

Once the granulometric transforms are able to disperse information of size and shape in residual images, other different representations of the histogram of patterns of objects of interest can be generated, since an adequate amount of transforms is used. The number of necessary transforms, that is, the largest values of $k$, depends upon the application, because it is directly associated to the dimension of the feature vector to be used in the application of pattern recognition.

\subsection{Morphological Similarity Index} \label{subsec_morphindex}

From the definition of pattern spectrum and accumulated morphological distribution, it is possible to establish a \emph{morphological similarity index}, $Q_M(f,g,h)$, to express the similarity of image $f:S\rightarrow [0,1]$ i reference to image $g:S\rightarrow [0,1]$, given the structure element $h:S\rightarrow [0,1]$, as follows:
\begin{equation}
  Q_M(f,g,h)= \exp\left( - \left[ \frac { \sum_{k=0}^\infty (\xi_{f,h}[k]-\xi_{g,h}[k])^2 } { \sum_{k=0}^\infty \xi_{g,h}^2[k] } \right]^{1/2} \right),
\end{equation}
where $\xi_{f,h}[k]$ and $\xi_{g,h}[k]$ are the values of the pattern spectra of images $f$ and $g$ given the structure element $h$ in the $k$-th order, respectively. For the case of binary images, there exists $k_{\max}$ for
$$
\Xi_{f,h}[k_{\max}] \approx \Xi_{f,h}[k_{\max}-1]
$$
and
$$
\Xi_{g,h}[k_{\max}] \approx \Xi_{g,h}[k_{\max}-1].
$$

Consequently, the expression can be reduced to the following computable expression:
\begin{equation}
  Q_M(f,g,h)= \exp\left( - \left[ \frac { \sum_{k=0}^{k_{\max}} (\xi_{f,h}[k]-\xi_{g,h}[k])^2 } { \sum_{k=0}^{k_{\max}} \xi_{g,h}^2[k] } \right]^{1/2} \right).
\end{equation}

Although pattern spectra are unique for binary images, morphological similarity indexes can also be used in gray-level images as measurements of morphological similarity and textures.

\section{Results} \label{sec_results}

To evaluate objectively the classification results, there were used three methods: the index $\kappa$, the \emph{overall accuracy} and the \emph{confusion matrix}. The subjective evaluation was performed by the specialist knowledge of a pathologist. Image background ($C_3$), gray and white matter ($C_2$) and cerebrospinal fluid ($C_1$) were associated to the colors white, gray and black, respectively.

\begin{figure}
    \centering
        \includegraphics[width=0.45\textwidth]{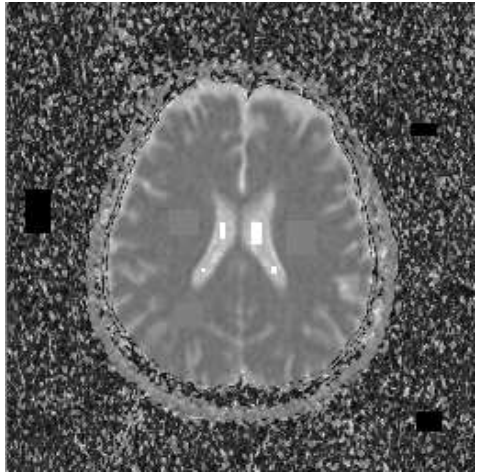}
    \caption{Training set}
    \label{fig:quadro_rois_treino}
\end{figure}

\begin{figure}
    \centering
        \includegraphics[width=0.45\textwidth]{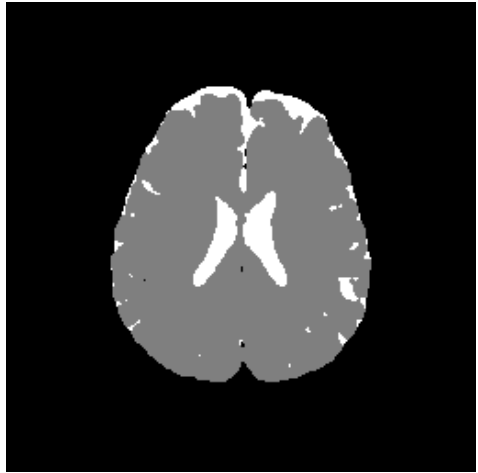}
    \caption{Ground truth image (13th slice) generated by polynomial net classification}
    \label{fig:classPO_13}
\end{figure}

\begin{figure}
    \centering
        \includegraphics[width=0.45\textwidth]{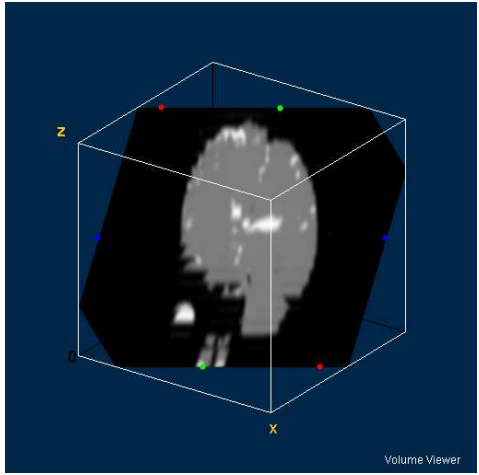}
    \caption{Ground truth volume generated by polynomial net classification}
    \label{fig:VistaVolumePO}
\end{figure}

Figure \ref{fig:quadro_rois_treino} shows the training set mounted on the 13th slice of the volume of ADC maps. Figures \ref{fig:classPO_13} and \ref{fig:VistaVolumePO} show the ground truth volume and the 13th slice, respectively.

The \emph{confusion matrix} for a universe of classes of interest $\Omega=\{C_1,C_2,\dots,C_m\}$ is a $m\times m$ matrix $\mathbf{T}=[t_{i,j}]_{m\times m}$ where each element $t_{i,j}$ represents the number of objects belonging to class $C_j$ but classified as $C_i$ \cite{duda2001,landgrebe2002}.

The \emph{overall accuracy} $\phi$ is the rate between the number of objects correctly classified and the total number of objects, defined as follows \cite{duda2001,landgrebe2002}:
\begin{equation}
  \phi=\rho_v=\frac{\sum_{i=1}^m t_{i,i}} {\sum_{i=1}^m \sum_{j=1}^m t_{i,j}}.
\end{equation}

The \emph{index} $\kappa$ is an statistical correlation rate defined as follows \cite{duda2001}:
\begin{equation}
  \kappa=\frac{\rho_v-\rho_z} {1-\rho_z},
\end{equation}
where
\begin{equation}
  \rho_z=\frac {\sum_{i=1}^m (\sum_{j=1}^m t_{i,j})(\sum_{j=1}^m t_{j,i})} {(\sum_{i=1}^m \sum_{j=1}^m t_{i,j})^2}.
\end{equation}

For the task of classification we assume that the classes of interest are separable by hyperquadrics. Hence a 2-degree polynomial network was chosen to classify the original volume and generate the ground truth volume. The polynomial degree was empirically determined by its gradual increasing until the differences between the present classification and the immediate previous reach a value less than a determined threshold. Such differences were evaluated using Wang's fidelity index \cite{wang2002}, given by the following expression:
\begin{equation}
  Q_W(f,g)=\frac {4\mu_f\mu_g\sigma_{f,g}} {(\mu_f^2+\mu_g^2)(\sigma_f^2+\sigma_g^2)},
\end{equation}
where $f$ and $g$ are monospectral images to be compared, and $\mu_f$, $\mu_g$, $\sigma_f^2$, $\sigma_g^2$, and $\sigma_{f,g}$ are the sample average of $f$, the sample average of $g$, the sample variance of $f$, the sample variance of $g$, and the correlation between $f$ and $g$, respectively.

Consequently, the measured similarities between the results obtained by the use of 4-degree and 3-degree polynomial classification related to the 2-degree classification were 0.9816 and 0.9904, respectively. Therefore, we have chosen the result obtained by the 2-degree polynomial classification.

The polynomial net is a 2-layer network, where the first layer is a multiplicative net responsible for the generation of all 2-degree polynomial terms associated to 3 inputs; the second layer is an one-layered perceptron with initial learning rate $\eta_0=0.1$, training error $\epsilon=0.05$, and maximum of 200 training iterations. The second layer is responsible for the calculation of the coefficients of the polynomial dedicated to model the discriminant function of each class of interest \cite{duda2001,sklansky1981}. Hence, the polynomial net is a polynomial approximator. Learning rate and training error were empirically defined.

\begin{figure}
    \centering
        \includegraphics[width=0.45\textwidth]{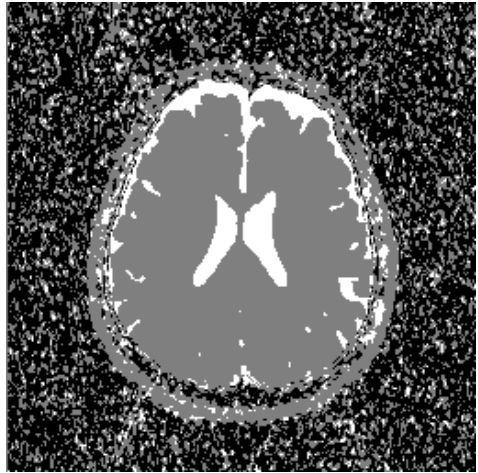}
    \caption{Fuzzy c-means classification of 13th slice using the ADC map}
    \label{fig:classCM_ADC_13}
\end{figure}
\begin{figure}
    \centering
        \includegraphics[width=0.45\textwidth]{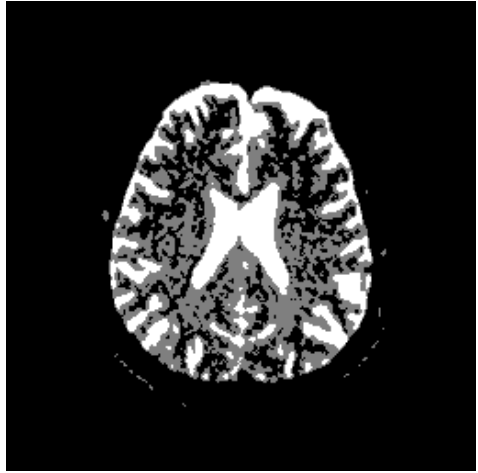}
    \caption{Multilayer perceptron classification of 13th slice}
    \label{fig:classRNP_13}
\end{figure}
\begin{figure}
    \centering
        \includegraphics[width=0.45\textwidth]{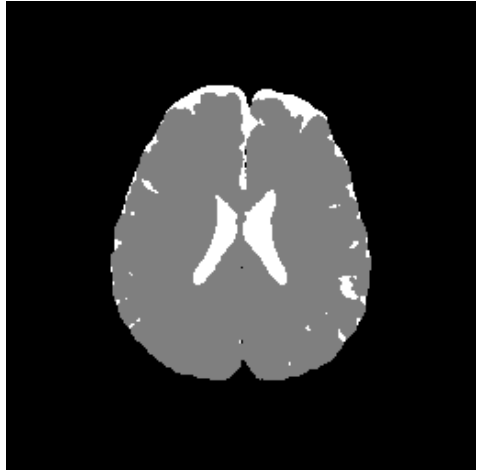}
    \caption{Radial basis function network classification of 13th slice}
    \label{fig:classRBF_13}
\end{figure}
\begin{figure}
    \centering
        \includegraphics[width=0.45\textwidth]{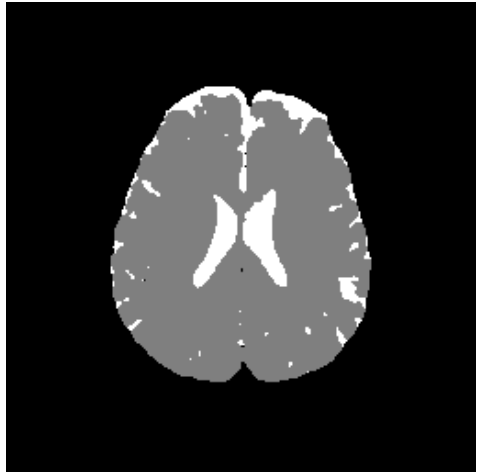}
    \caption{Kohonen self-organized map classification of 13th slice}
    \label{fig:classKO_13}
\end{figure}
\begin{figure}
    \centering
        \includegraphics[width=0.45\textwidth]{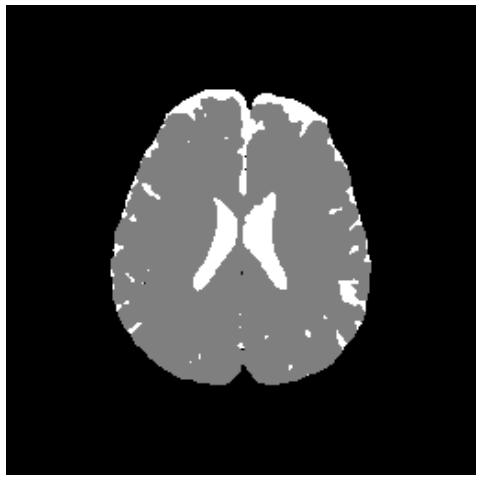}
    \caption{Kohonen LVQ classification of 13th slice}
    \label{fig:classLVQ_13}
\end{figure}
\begin{figure}
    \centering
        \includegraphics[width=0.45\textwidth]{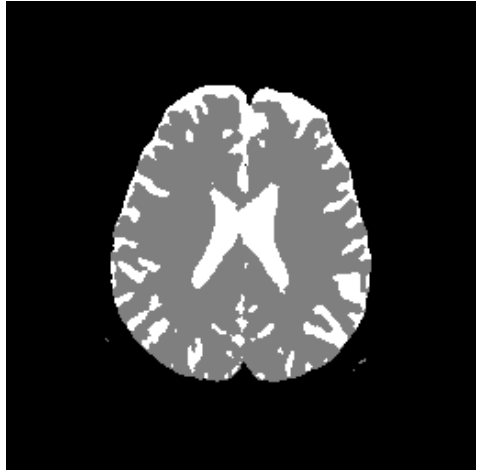}
    \caption{Fuzzy c-means classification of 13th slice}
    \label{fig:classCM_13}
\end{figure}
\begin{figure}
    \centering
        \includegraphics[width=0.45\textwidth]{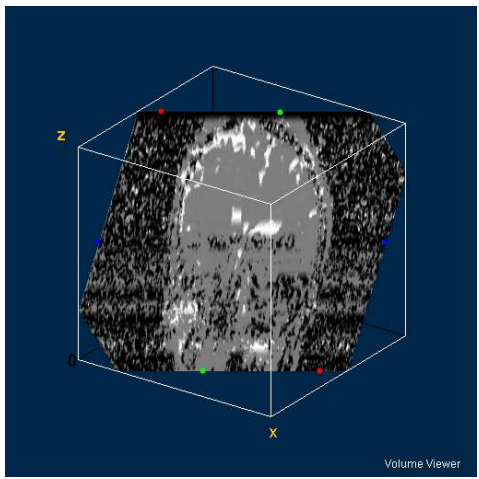}
    \caption{Fuzzy c-means classification volume using the ADC map}
    \label{fig:VistaVolumeCM_ADC}
\end{figure}
\begin{figure}
    \centering
        \includegraphics[width=0.45\textwidth]{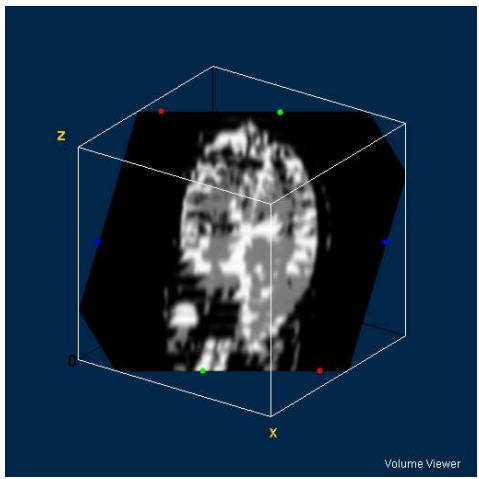}
    \caption{Multilayer perceptron classification volume}
    \label{fig:VistaVolumeRNP}
\end{figure}
\begin{figure}
    \centering
        \includegraphics[width=0.45\textwidth]{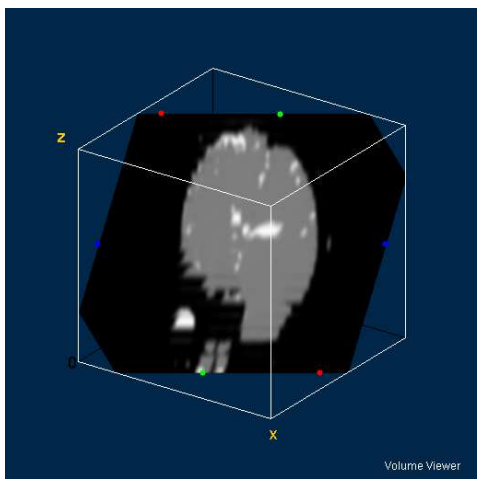}
    \caption{Radial basis function network classification volume}
    \label{fig:VistaVolumeRBF}
\end{figure}
\begin{figure}
    \centering
        \includegraphics[width=0.45\textwidth]{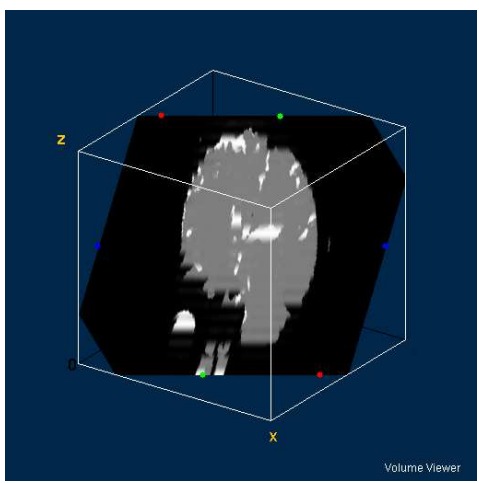}
    \caption{Kohonen self-organized network classification volume}
    \label{fig:VistaVolumeKO}
\end{figure}
\begin{figure}
    \centering
        \includegraphics[width=0.45\textwidth]{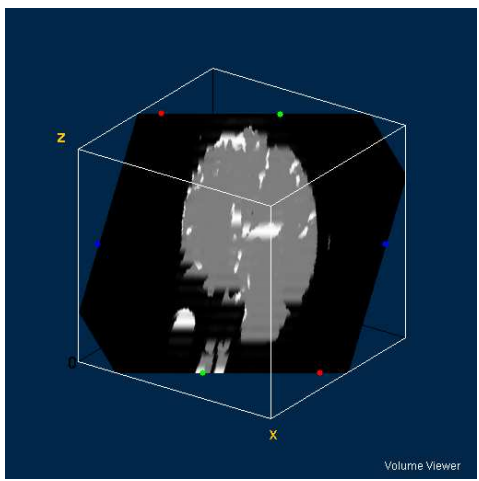}
    \caption{Kohonen LVQ classification volume}
    \label{fig:VistaVolumeLVQ}
\end{figure}
\begin{figure}
    \centering
        \includegraphics[width=0.45\textwidth]{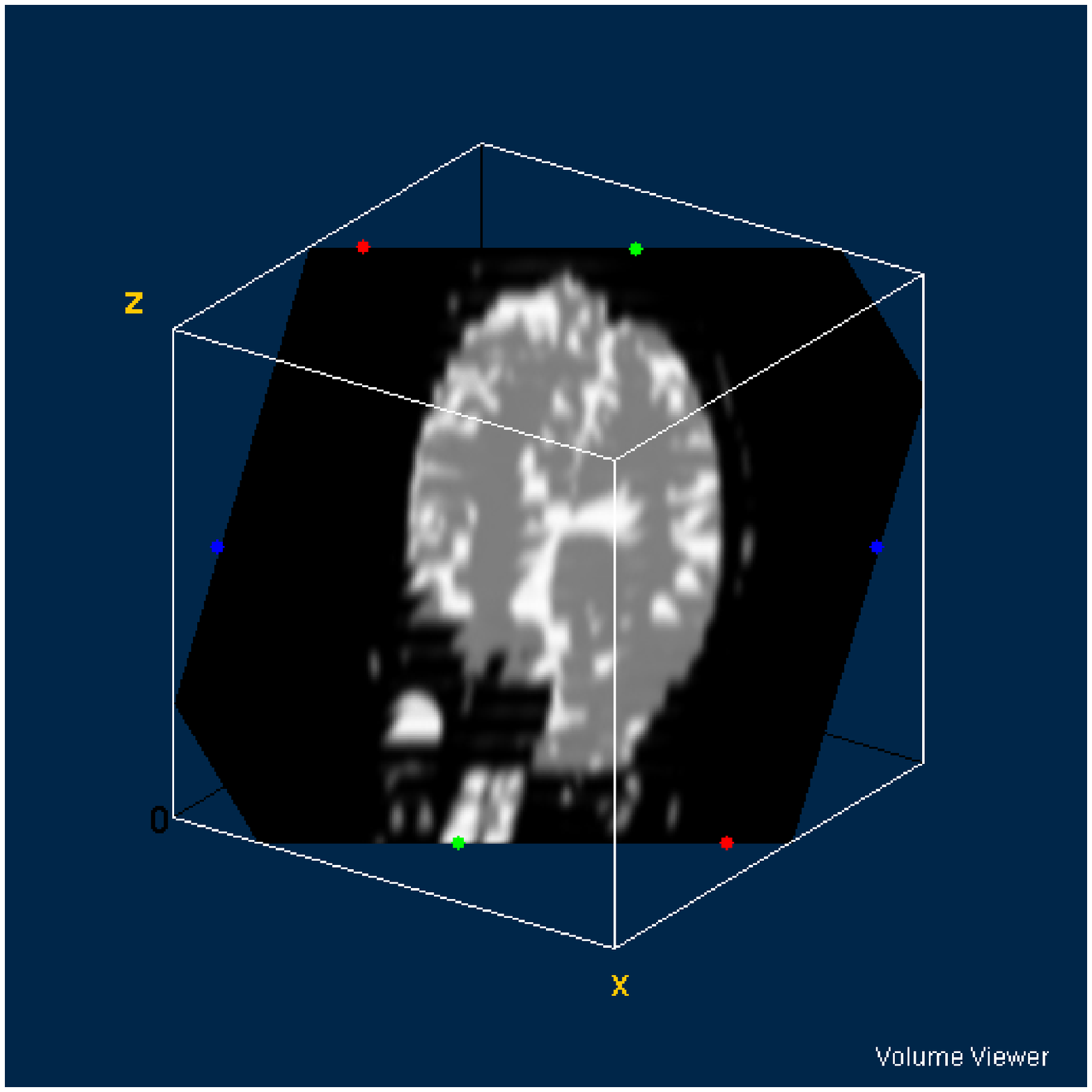}
    \caption{Fuzzy c-means classification volume}
    \label{fig:VistaVolumeCM}
\end{figure}

Figure \ref{fig:VistaVolumeCM_ADC} shows the classification of the ADC map using fuzzy c-means classifier. Figures \ref{fig:VistaVolumeRNP}, \ref{fig:VistaVolumeRBF}, \ref{fig:VistaVolumeKO}, \ref{fig:VistaVolumeLVQ} and \ref{fig:VistaVolumeCM} show the classification of the volume of multispetral synthetic images composed by the volumes shown in figures \ref{fig:VistaVolumet1}, \ref{fig:VistaVolumet2} and \ref{fig:VistaVolumet3}, using methods MLP, RBF, KO, LVQ and CM, respectively. Table \ref{tab:ResultadosClassificacao} presents index $\kappa$ and overall accuracy $\phi$, whilst table \ref{tab:AreasClassificacao} explicits percentual volumes $V_1$, $V_2$ and $V_3$ occupied by the classes of interest $C_1$, $C_2$ and $C_3$, respectively, as well as the ration between the volume of cerebrospinal fluid and the total volume of gray and white matter, namely the fluid-matter ratio, $V_1/V_2$. Figure \ref{fig:classCM_ADC_13} presents the 13th slice of classification volume of ADC maps using method CM, whilst figures \ref{fig:classRNP_13}, \ref{fig:classRBF_13}, \ref{fig:classKO_13}, \ref{fig:classLVQ_13} and \ref{fig:classCM_13} show 13th slices of the 20-slice volume generated by the classification according to methods MLP, RBF, KO, LVQ and CM, in this order. 

\begin{figure}
    \centering
        \includegraphics[width=0.45\textwidth]{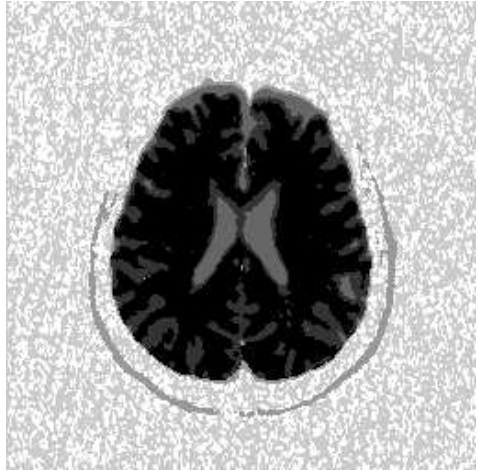}
    \caption{13th slice of objective dialectical classification}
    \label{fig:classDLT6c_13}
\end{figure}

\begin{figure}
    \centering
        \includegraphics[width=0.45\textwidth]{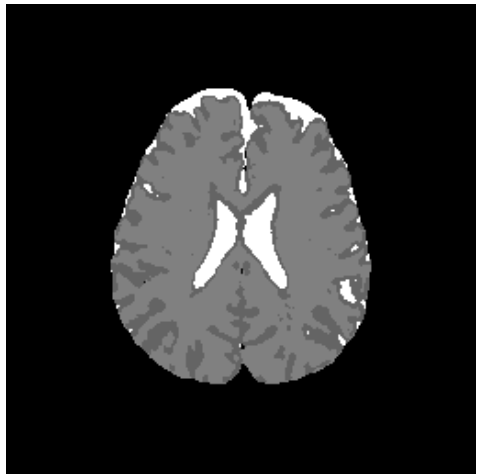}
    \caption{13th slice of objective dialectical classification after post-labeling}
    \label{fig:classDLT_13}
\end{figure}

\begin{figure}
    \centering
        \includegraphics[width=0.45\textwidth]{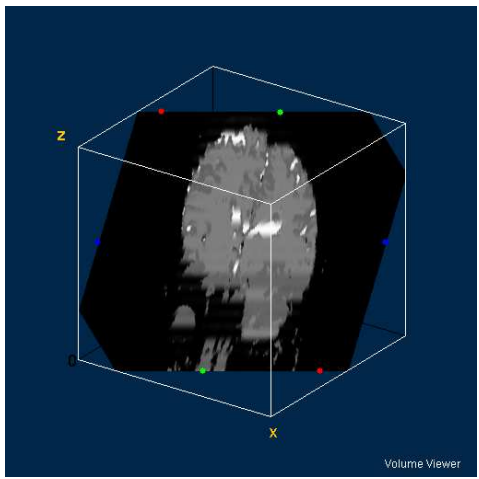}
    \caption{Objective dialectical classification volume after post-labeling}
    \label{fig:VistaVolumeDLT}
\end{figure}

Figure \ref{fig:classDLT6c_13} shows 13th slice of classification result obtained by the use of the objective dialectical classifier. Figure \ref{fig:classDLT_13} exhibits 13th slice of classification result after performing post-labeling, whilst figure \ref{fig:VistaVolumeDLT} shows the entire volume generated by ODC classification.

The training process of ODC algorithm resulted in 6 classes. These classes were reduced to 4 after manual post-labeling, merging 3 classes out of brain region, namely image background, noise and cranial box. The post-labeling is manual because all 3 cited regions are statistically different and, consequently, they are merged due to our interest in classes more related to the brain regions. On figure \ref{fig:classDLT_13} it is possible to notice that ODC was able to distinguish white matter from the gray matter present in the interface between liquor and white matter.

\begin{table}
    \caption{Overall accuracy $\phi$ (\%) and index $\kappa$ related to the classification methods}
  \begin{center}
  \begin{tabular} {ccccccc}
  \hline
  {} & {MLP} & {RBF} & {KO} & {LVQ} & {CM} & {ADC-CM}\\
  \hline
  {$\phi$ (\%)} & {88.5420} & {99.3587}  & {99.3647} & {99.3683} & {93.9364} & {58.1154}\\
  {$\kappa$} & {0.6081} & {0.9681}  & {0.9688} & {0.9689} & {0.7755} & {0.2495}\\
  \hline
  \end{tabular}
  \end{center}
    \label{tab:ResultadosClassificacao}
\end{table}

\begin{table}
    \caption{Percentual volumes and fluid-matter ratio related to the classification methods}
  \begin{center}
  \begin{tabular} {cccccccc}
  \hline
  {} & {MLP} & {RBF} & {KO} & {LVQ} & {CM} & {ADC-CM} & {PO}\\
  \hline
  {$V_1$ (\%)} & {7.607} & {1.612} & {2.186} & {2.185} & {6.262} & {18.743} & {1.697}\\
  {$V_2$ (\%)} & {11.546} & {17.187} & {16.398} & {16.394} & {13.922} & {34.354} & {17.010}\\
  {$V_3$ (\%)} & {80.847} & {81.201} & {81.416} & {81.421} & {79.816} & {46.903} & {81.293}\\
  {$V_1/V_2$} & {0.659} & {0.094} & {0.133} & {0.133} & {0.450} & {0.546} & {0.100}\\
  \hline
  \end{tabular}
  \end{center}
    \label{tab:AreasClassificacao}
\end{table}

\begin{table}
    \caption{Generalization indexes for the classification methods related to all bands of images}
  \begin{center}
  \begin{tabular} {ccccccc}
  \hline
  {} & {MLP} & {RBF} & {KO} & {LVQ} & {CM} & {ADC-CM}\\
  \hline
  {$\Delta \kappa/\bar{\kappa}$} & {0.7398} & {0.9623} & {0.9634} & {0.9635} & {0.8295} & {0.4519}\\
  {$\Delta \phi/\bar{\phi}$} & {0.9560} & {0.9956} & {0.9979} & {0.9979} & {0.9715} & {0.8577}\\
  \hline
  \end{tabular}
  \end{center}
    \label{tab:IndiceGeneralizacao}
\end{table}

\begin{figure}
	\centering
		\includegraphics[width=0.90\textwidth]{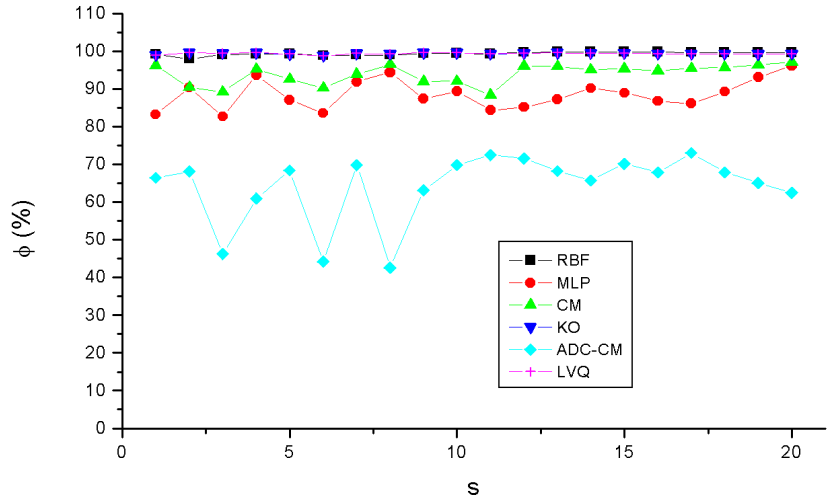}
	\caption{Behavior of overall accuracy $\phi$ related to $s$-th slice}
	\label{fig:Graph_phi}
\end{figure}

\begin{figure}
	\centering
		\includegraphics[width=0.90\textwidth]{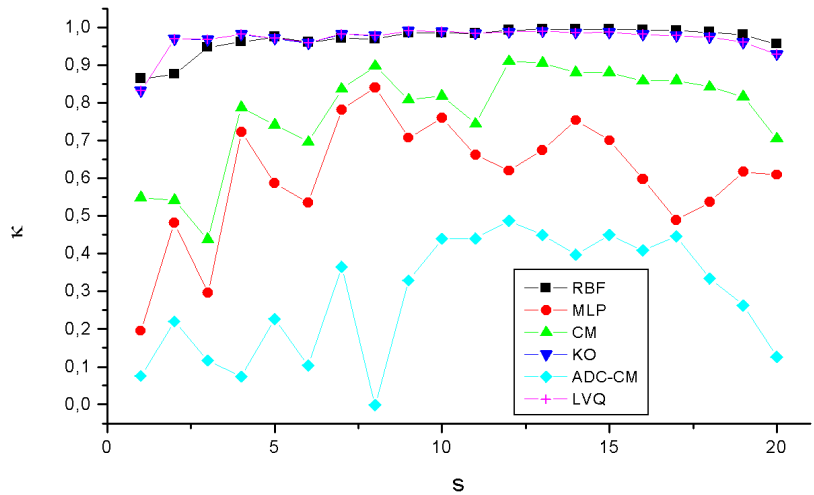}
	\caption{Behavior of index $\kappa$ related to $s$-th slice}
	\label{fig:Graph_kappa}
\end{figure}

The training of the used classification methods was performed using 13th slice. The resultant parameters of the classifiers were generalized to the other 19 slices. Figures \ref{fig:Graph_phi} and \ref{fig:Graph_kappa} illustrate the behavior of indexes $\phi$ and $\kappa$ according to $s$-th slice, where $1\leq s\leq 20$. From these graphics it is possible to study the \emph{generalization index} of each method, here defined as $\Delta \kappa/\bar{\kappa}$, resulting the values on table \ref{tab:IndiceGeneralizacao}, where we can notice that the index of generalization of MLP method is relatively low, 0.7398, if compared to the others, varying from 0.8295 until 0.9635. However, method ADC-CM got the lowest index of generalization, 0.4519.

From table \ref{tab:IndiceGeneralizacao} we can make the conclusion that it is not possible to define the index of generalization as $\Delta \phi/\bar{\phi}$, because the lowest value, despite the fact that it was measured from the result of method ADC-CM, amounts to 0.8577 and, consequently, does not correspond to reality. This assumption can be confirmed in case we take a look at its statistics $\kappa$ of 0.4519.

\begin{figure}
	\centering
		\includegraphics[width=0.45\textwidth]{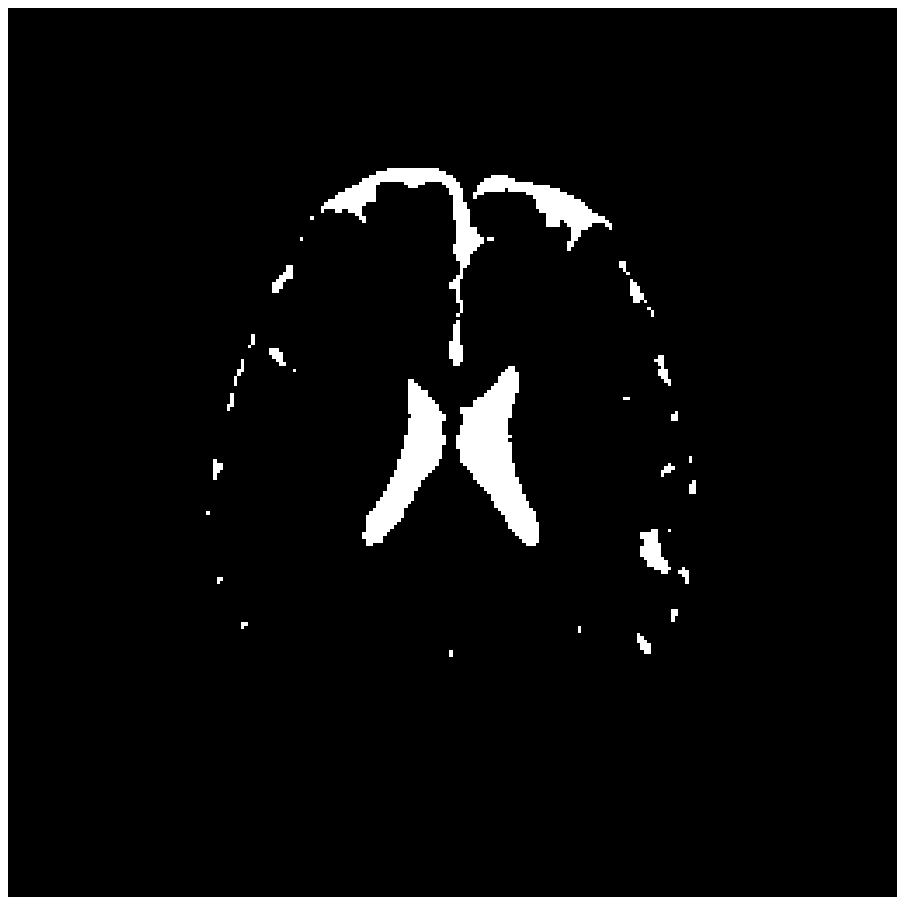}
	\caption{13th slice of liquor volume detected by method ODC}
	\label{fig:csfDLT_13}
\end{figure}

\begin{figure}
	\centering
		\includegraphics[width=0.45\textwidth]{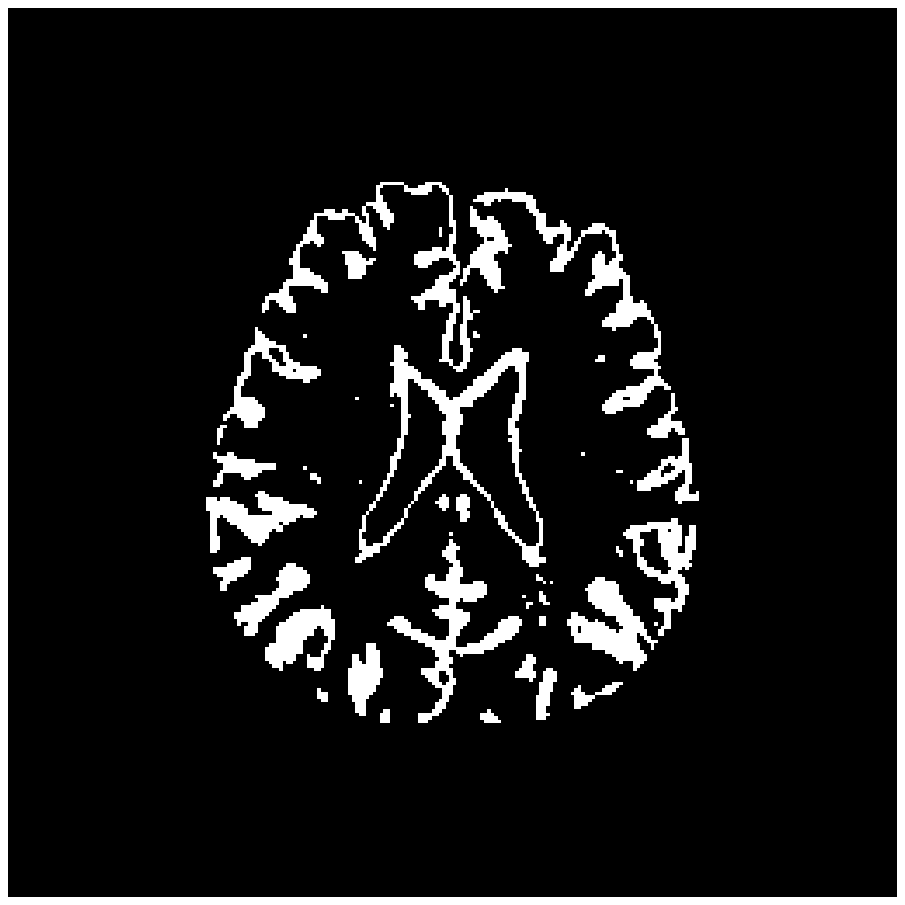}
	\caption{13th slice of gray matter volume detected by method ODC}
	\label{fig:gmDLT_13}
\end{figure}

\begin{figure}
	\centering
		\includegraphics[width=0.45\textwidth]{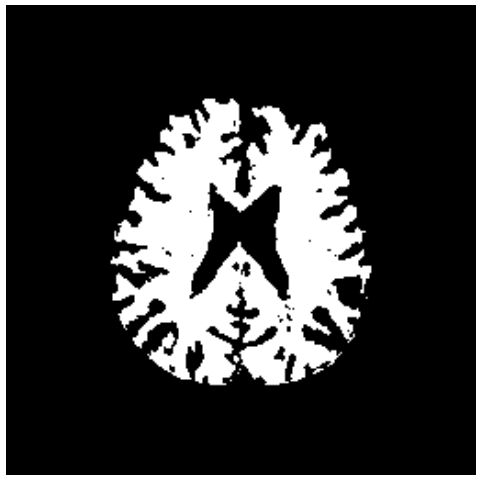}
	\caption{13th slice of white matter volume detected by method ODC}
	\label{fig:wmDLT_13}
\end{figure}

\begin{figure}
	\centering
		\includegraphics[width=0.90\textwidth]{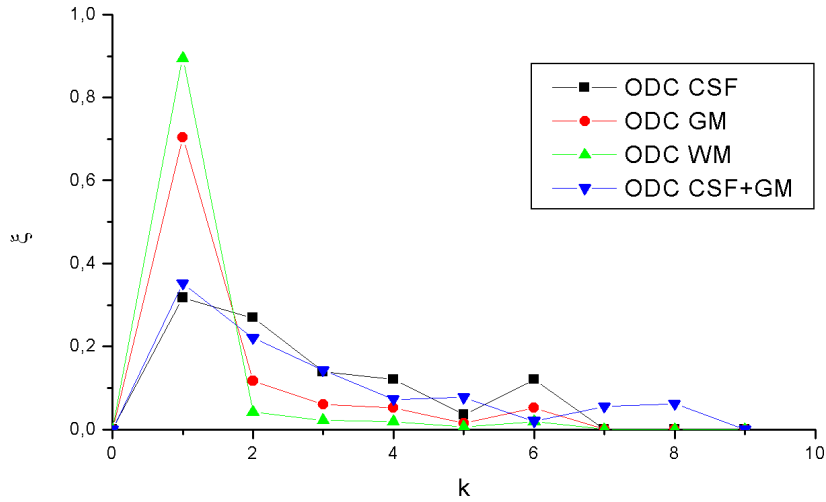}
	\caption{Morphological spectra of 13th slice of liquor, white matter and gray matter volume detected by method ODC}
	\label{fig:GraphEspecMorfo_ODC_13}
\end{figure}

\begin{figure}
	\centering
		\includegraphics[width=0.90\textwidth]{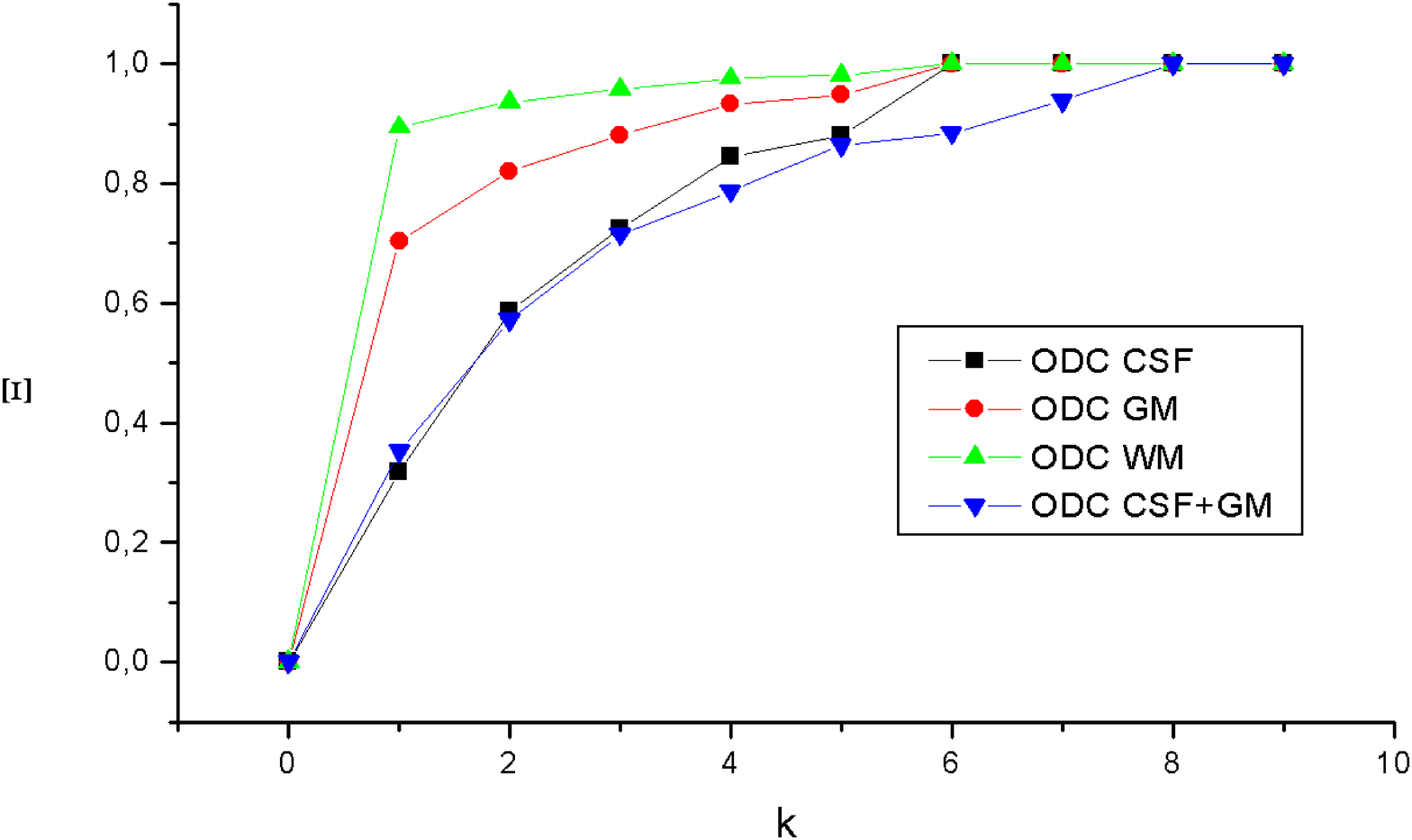}
	\caption{Accumulated morphological distributions of 13th slice of liquor, white matter and gray matter volume detected by method ODC}
	\label{fig:GraphDistMorfo_ODC_13}
\end{figure}

Figures \ref{fig:csfDLT_13}, \ref{fig:gmDLT_13} and \ref{fig:wmDLT_13} show binary images of 13th slice of ODC classification, showing liquor (\emph{cerebrospinal fluid}, CSF), gray matter (GM) and white matter (WM), respectively, whilst figures \ref{fig:GraphEspecMorfo_ODC_13} and \ref{fig:GraphDistMorfo_ODC_13} show the respective morphological spectra and accumulated morphological distributions. Both morphological spectra and accumulated morphological distributions were obtained using structure element square $3\times 3$ center $(2,2)$.
\begin{figure}
	\centering
		\includegraphics[width=0.45\textwidth]{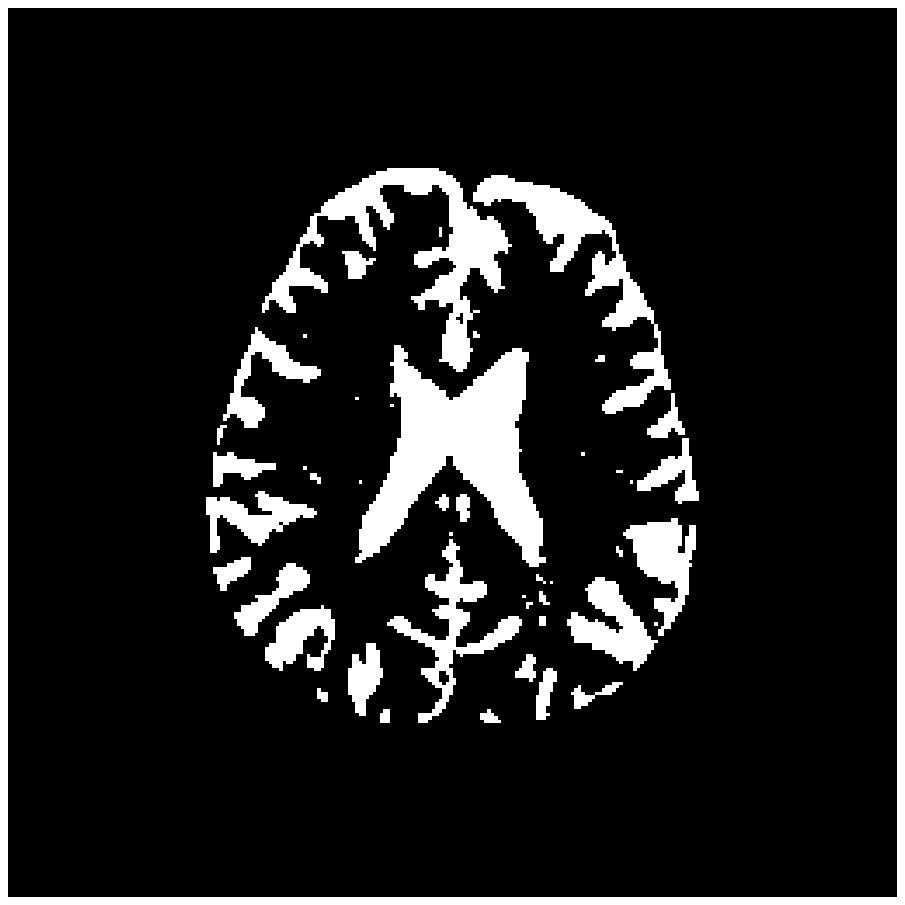}
	\caption{13th slice of liquor and gray matter volume detected by method ODC}
	\label{fig:csfgmDLT_13}
\end{figure}

Figure \ref{fig:csfgmDLT_13} illustrates liquor and gray matter (CSF+GM) of 13th slice detected by ODC method, whilst figures \ref{fig:GraphEspecMorfo_ODC_13} and \ref{fig:GraphDistMorfo_ODC_13} show the graphics of associated morphological spectrum and accumulated morphological distribution.

\begin{figure}
	\centering
		\includegraphics[width=0.45\textwidth]{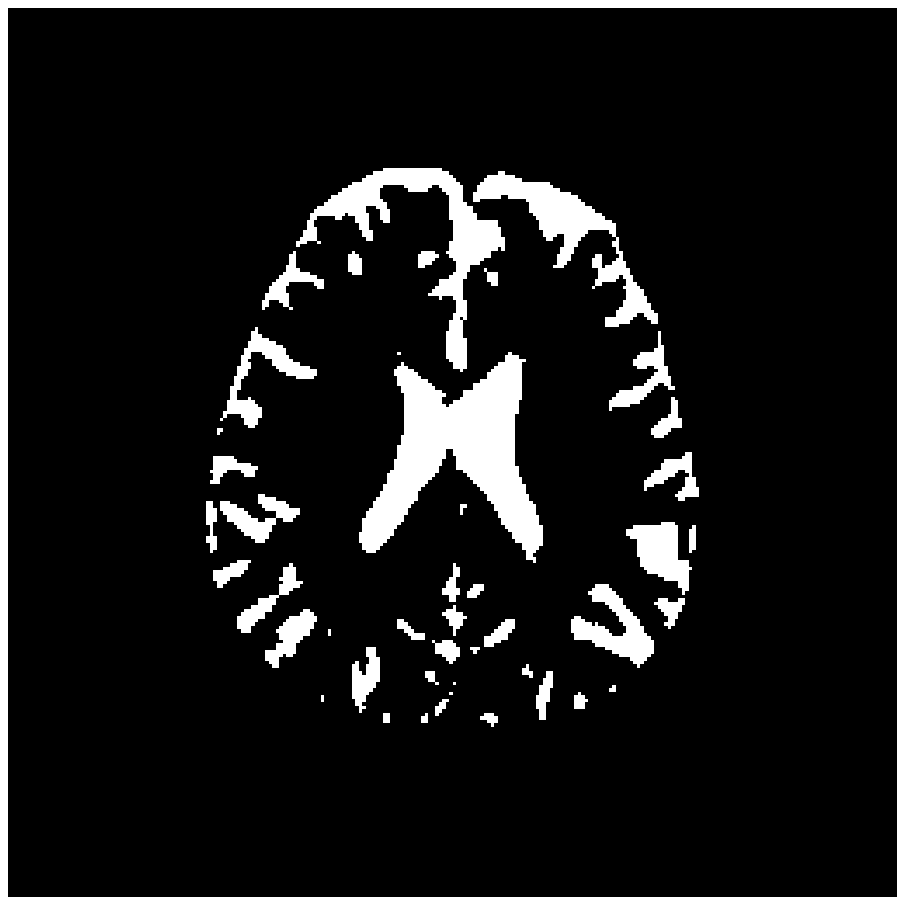}
	\caption{13th slice of the volume of liquor detected by method CM}
	\label{fig:csfCM_13}
\end{figure}

\begin{figure}
	\centering
		\includegraphics[width=0.45\textwidth]{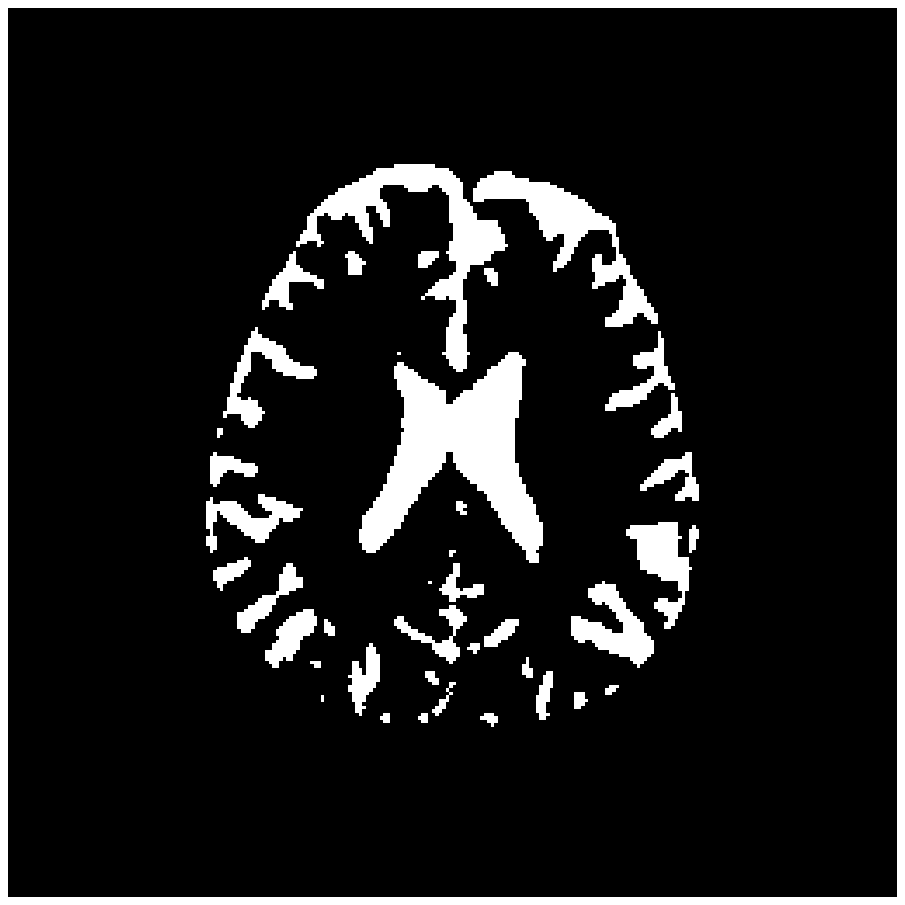}
	\caption{13th slice of the volume of liquor detected by method MLP}
	\label{fig:csfRNP_13}
\end{figure}

\begin{figure}
	\centering
		\includegraphics[width=0.90\textwidth]{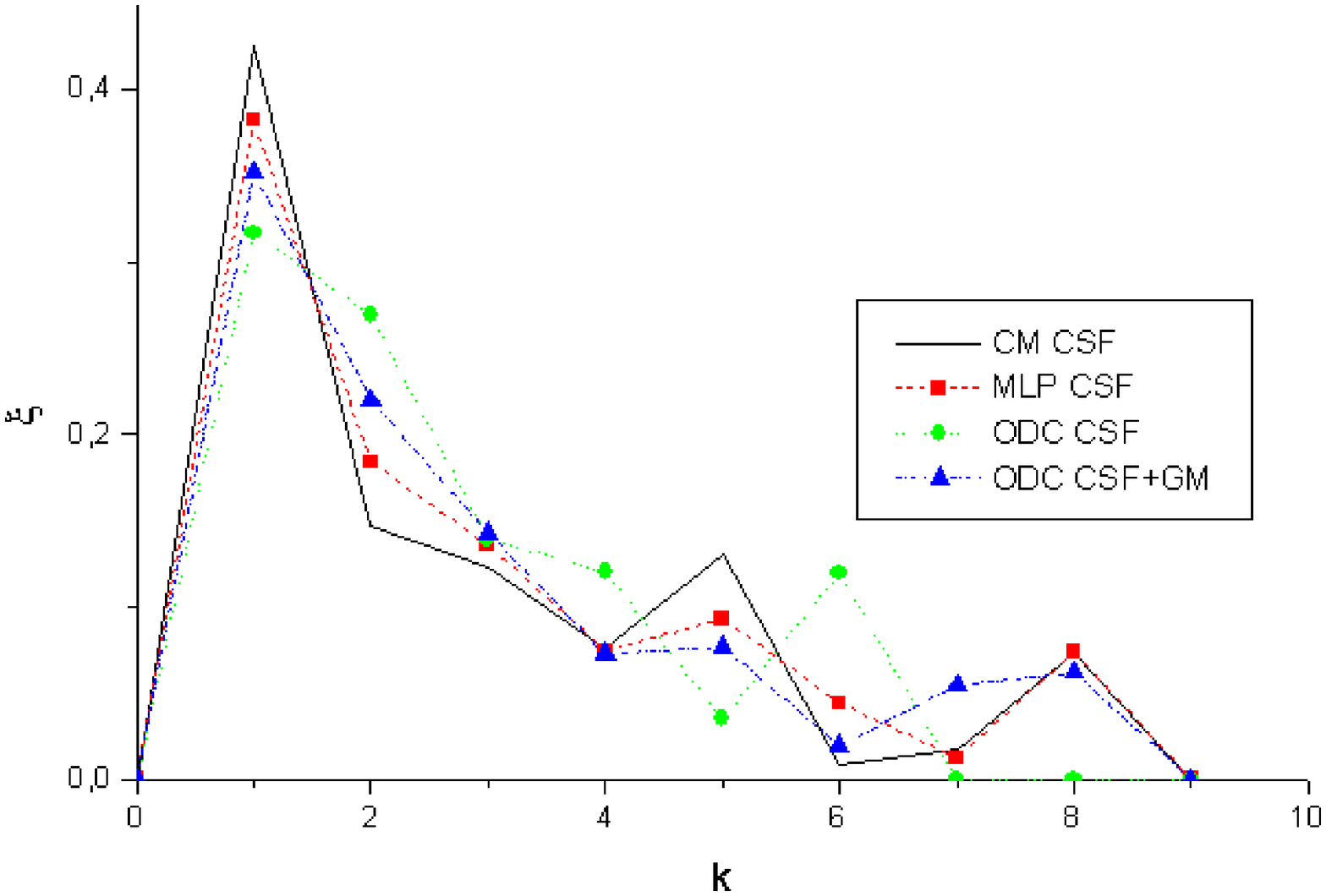}
	\caption{Morphological spectra of 13th slice of liquor volumes detected by methods CM, MLP and ODC, and of liquor and gray matter volume detected by method ODC}
	\label{fig:GraphEspecMorfo_CM_MLP_ODC_13}
\end{figure}

\begin{figure}
	\centering
		\includegraphics[width=0.90\textwidth]{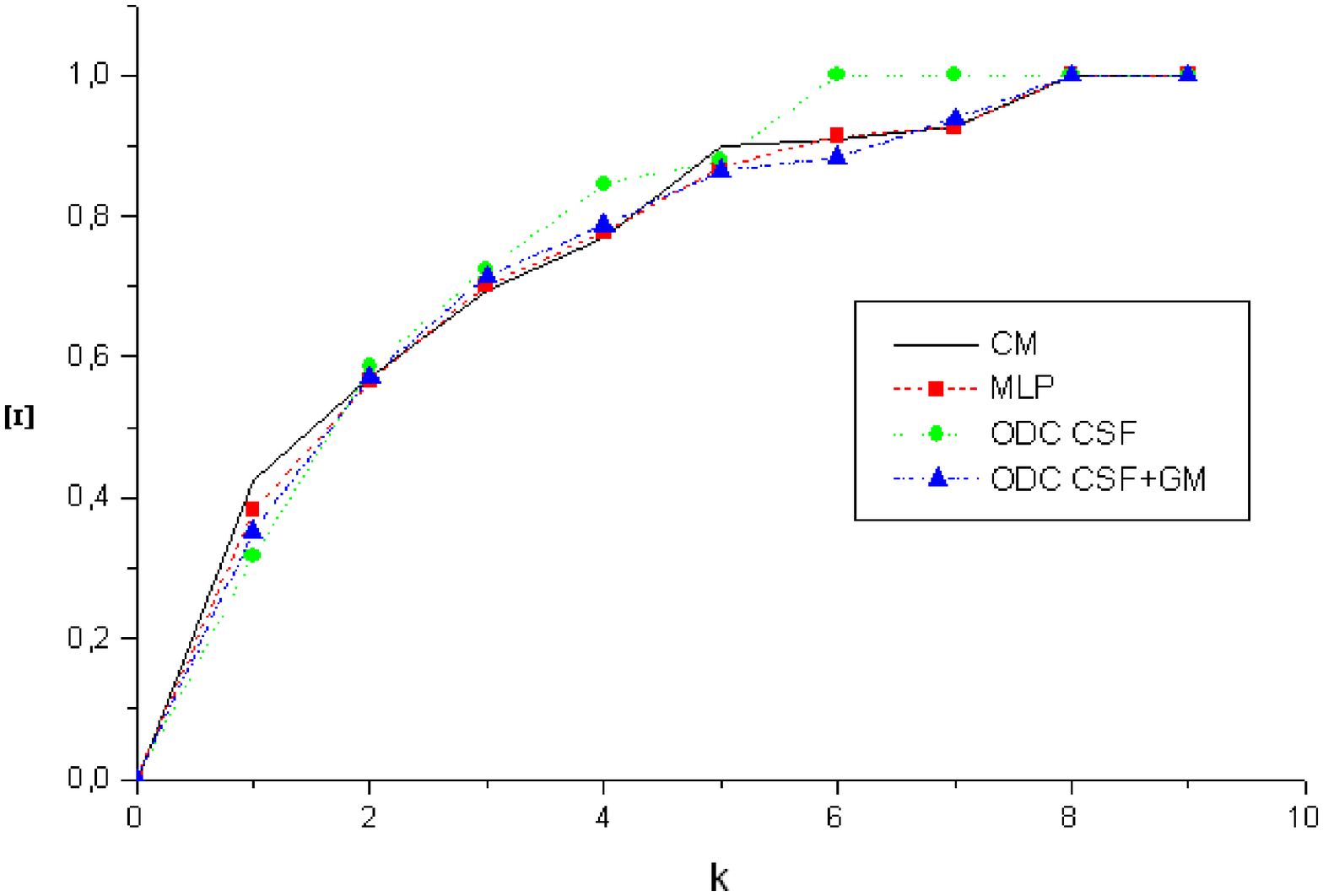}
	\caption{Accumulated morphological distributions of 13th slice of liquor volumes detected by methods CM, MLP and ODC, and of liquor and gray matter volume detected by method ODC}
	\label{fig:GraphDistMorfo_CM_MLP_ODC_13}
\end{figure}

Figures \ref{fig:csfCM_13} and \ref{fig:csfRNP_13} show 13th slice of the volume of liquor detected by CM and MLP, whilst figures \ref{fig:GraphEspecMorfo_CM_MLP_ODC_13} and \ref{fig:GraphDistMorfo_CM_MLP_ODC_13} show the respective morphological spectra and accumulated morphological distributions.

Comparing these results with those obtained from the volume of liquor detected by ODC method, it is possible to observe that methods CM and MLP associated liquor and gray matter to the same class liquor. It is confirmed by the observation of pattern spectra and values of morphological similarity index of the results of methods CM and MLP in reference to the result of ODC, namely 0.7274 and 0.8450, respectively.
\begin{figure}
	\centering
		\includegraphics[width=0.45\textwidth]{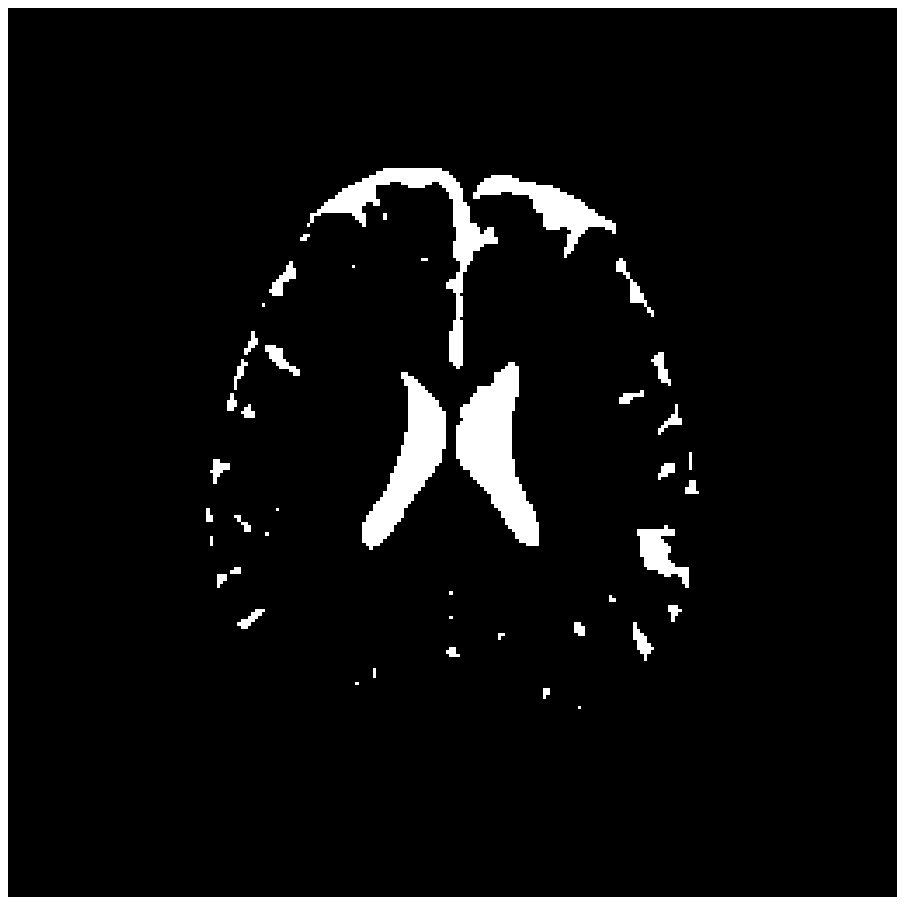}
	\caption{13th slice of the volume of liquor detected by method KO}
	\label{fig:csfKO_13}
\end{figure}
\begin{figure}
	\centering
		\includegraphics[width=0.45\textwidth]{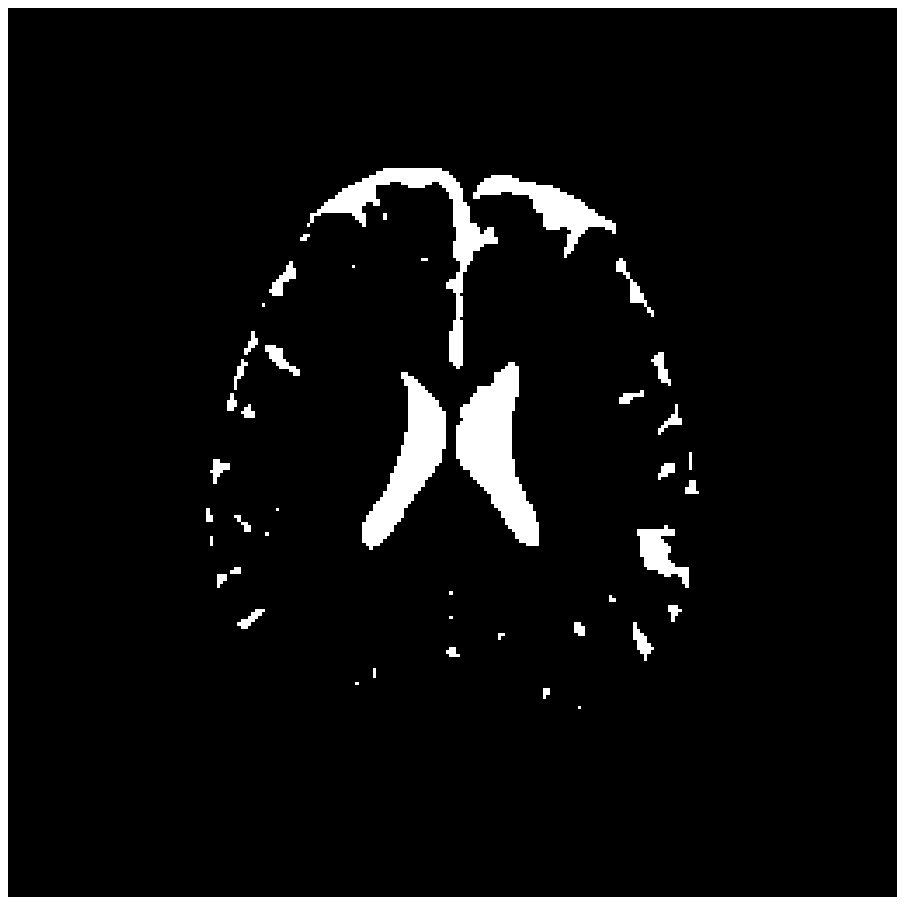}
	\caption{13th slice of the volume of liquor detected by method LVQ}
	\label{fig:csfLVQ_13}
\end{figure}
\begin{figure}
	\centering
		\includegraphics[width=0.45\textwidth]{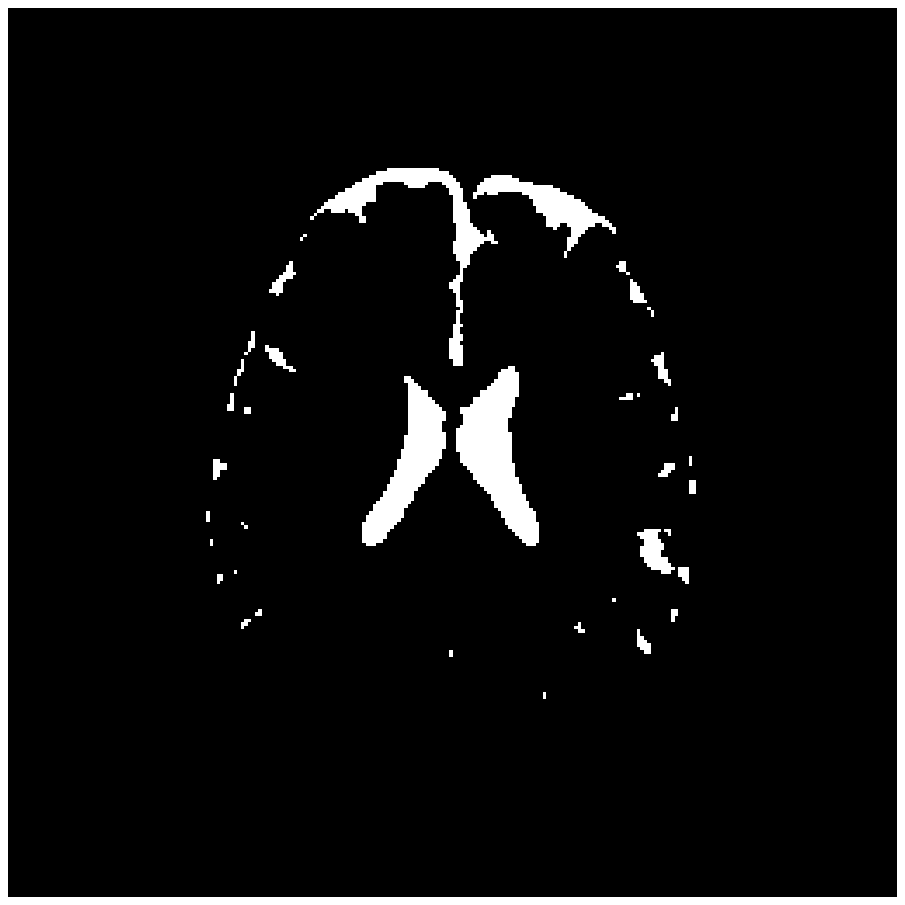}
	\caption{13th slice of the volume of liquor detected by method RBF}
	\label{fig:csfRBF_13}
\end{figure}

\begin{figure}
	\centering
		\includegraphics[width=0.45\textwidth]{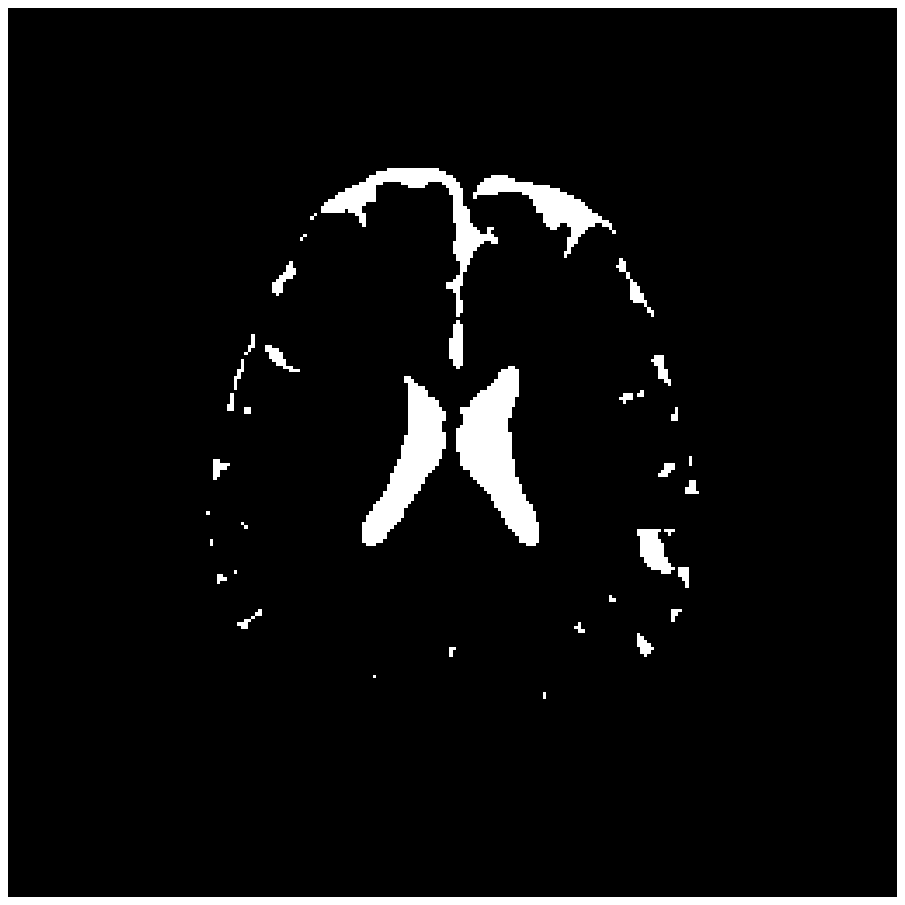}
	\caption{13th slice of the volume of liquor detected by method PO}
	\label{fig:csfPO_13}
\end{figure}

\begin{figure}
	\centering
		\includegraphics[width=0.90\textwidth]{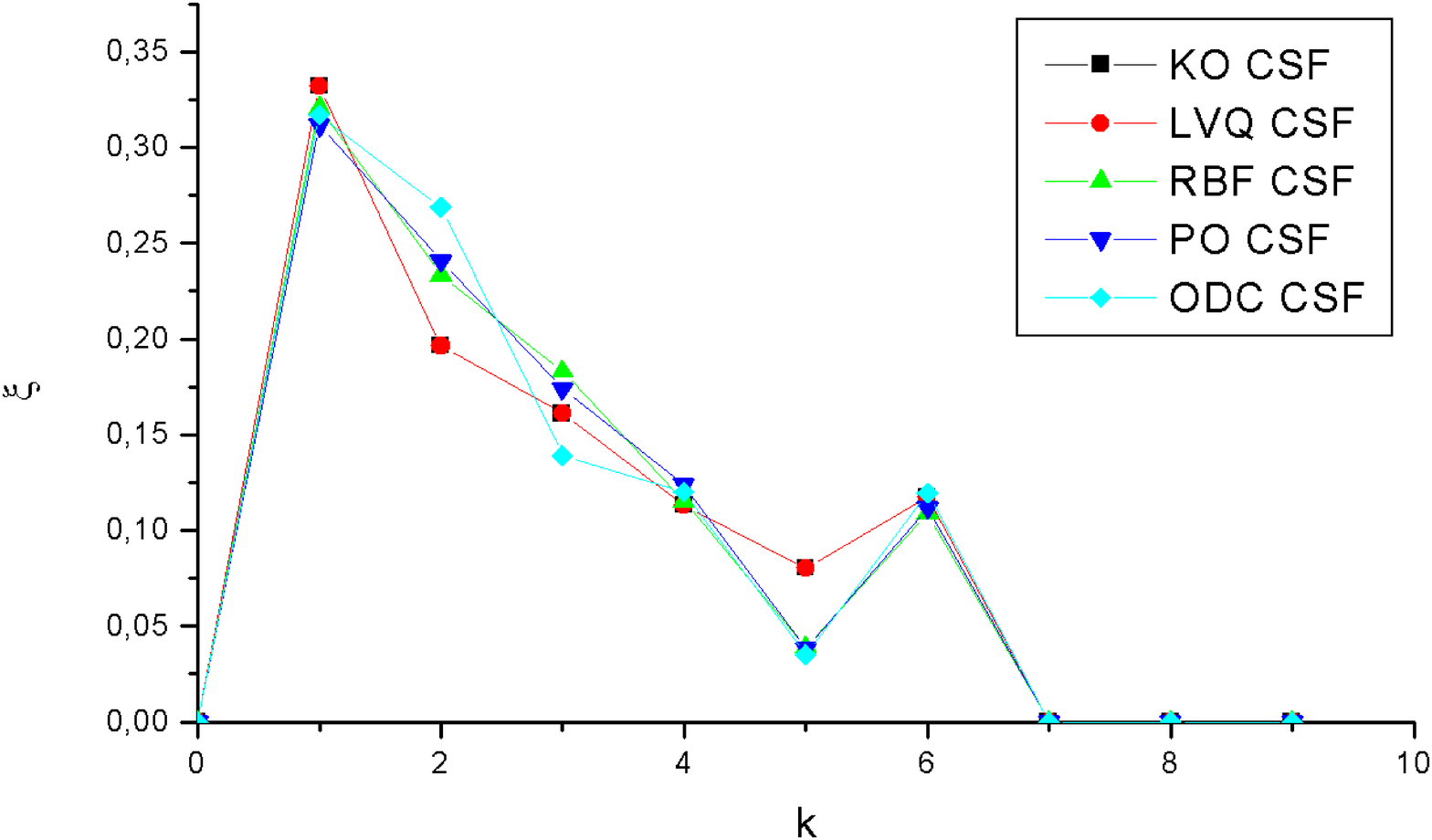}
	\caption{Morphological spectra of 13th slice of liquor volumes detected by methods KO, LVQ, RBF, PO and ODC, and of liquor and gray matter volume detected by method ODC}
	\label{fig:GraphEspecMorfo_KO_LVQ_RBF_PO_ODC_13}
\end{figure}

\begin{figure}
	\centering
		\includegraphics[width=0.90\textwidth]{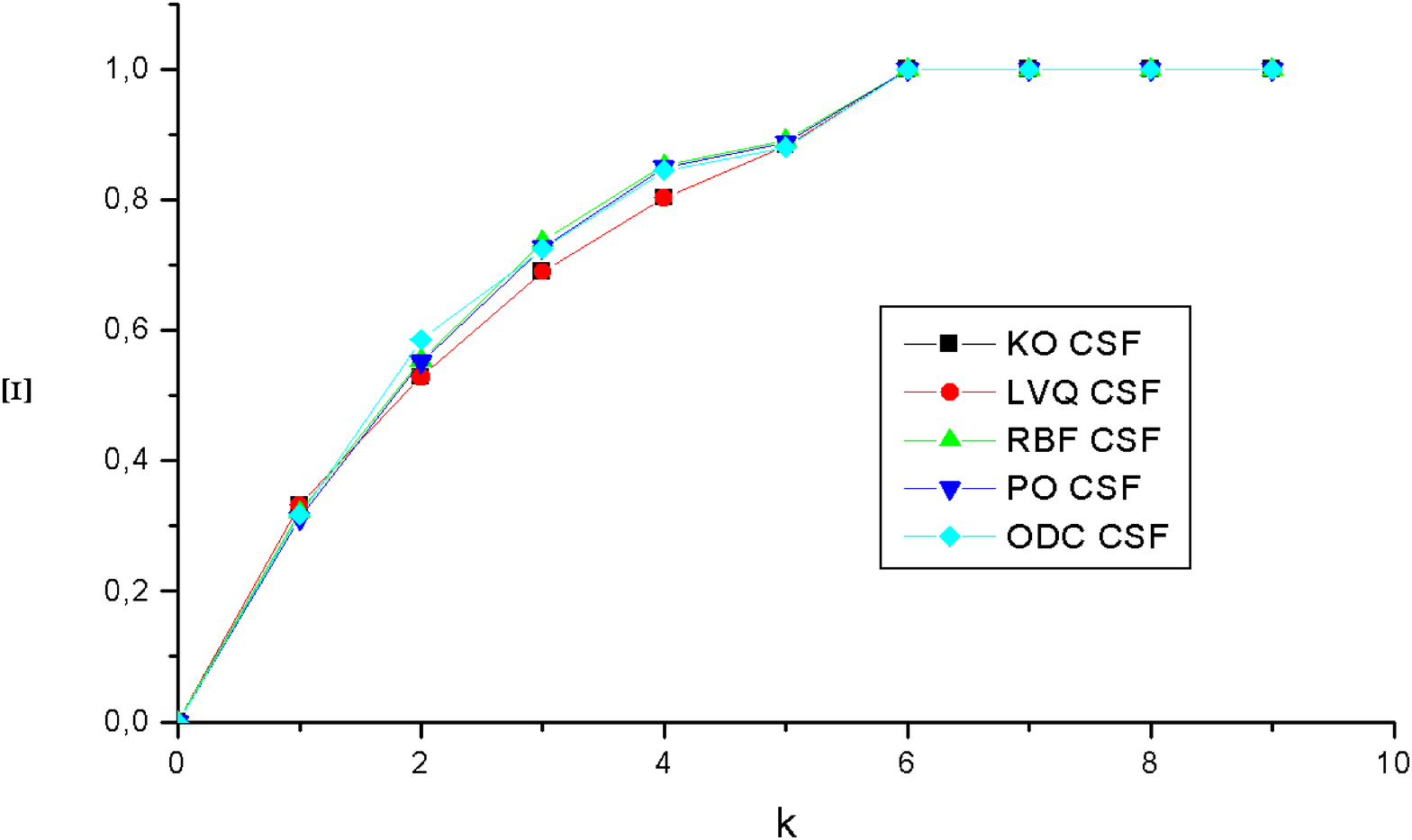}
	\caption{Accumulated morphological distributions of 13th slice of liquor volumes detected by methods KO, LVQ, RBF, PO and ODC, and of liquor and gray matter volume detected by method ODC}
	\label{fig:GraphDistMorfo_KO_LVQ_RBF_PO_ODC_13}
\end{figure}

Figures \ref{fig:csfKO_13}, \ref{fig:csfLVQ_13}, \ref{fig:csfRBF_13} and \ref{fig:csfPO_13} show 13th slice of liquor volume detected by methods KO, LVQ, RBF and PO, whilst figures \ref{fig:GraphEspecMorfo_KO_LVQ_RBF_PO_ODC_13} and \ref{fig:GraphDistMorfo_KO_LVQ_RBF_PO_ODC_13} exhibits their respective morphological spectra and accumulated morphological distributions.

Comparing these results to those obtained from the volume of liquor detected by method ODC, we can understand that methods KO, LVQ, RBF and ODC detected class liquor correctly. This fact can be confirmed by the morphological spectra and the values of morphological similarity index of results of methods KO, LVQ, RBF and ODC in reference to method PO, namely, 0.8554, 0.8554, 0.9619 and 0.9006, respectively.

\begin{figure}
    \centering
        \includegraphics[width=0.45\textwidth]{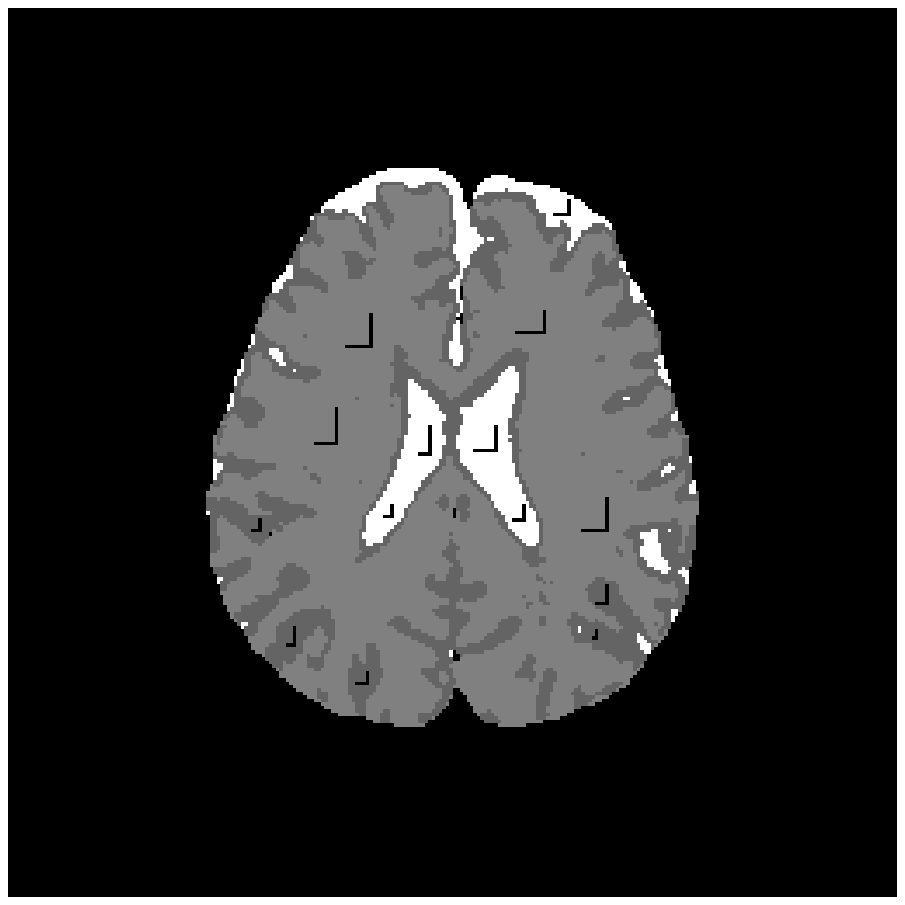}
    \caption{Training set on ODC result}
    \label{fig:quadro_rois_treino4c}
\end{figure}

\begin{figure}
	\centering
		\includegraphics[width=0.45\textwidth]{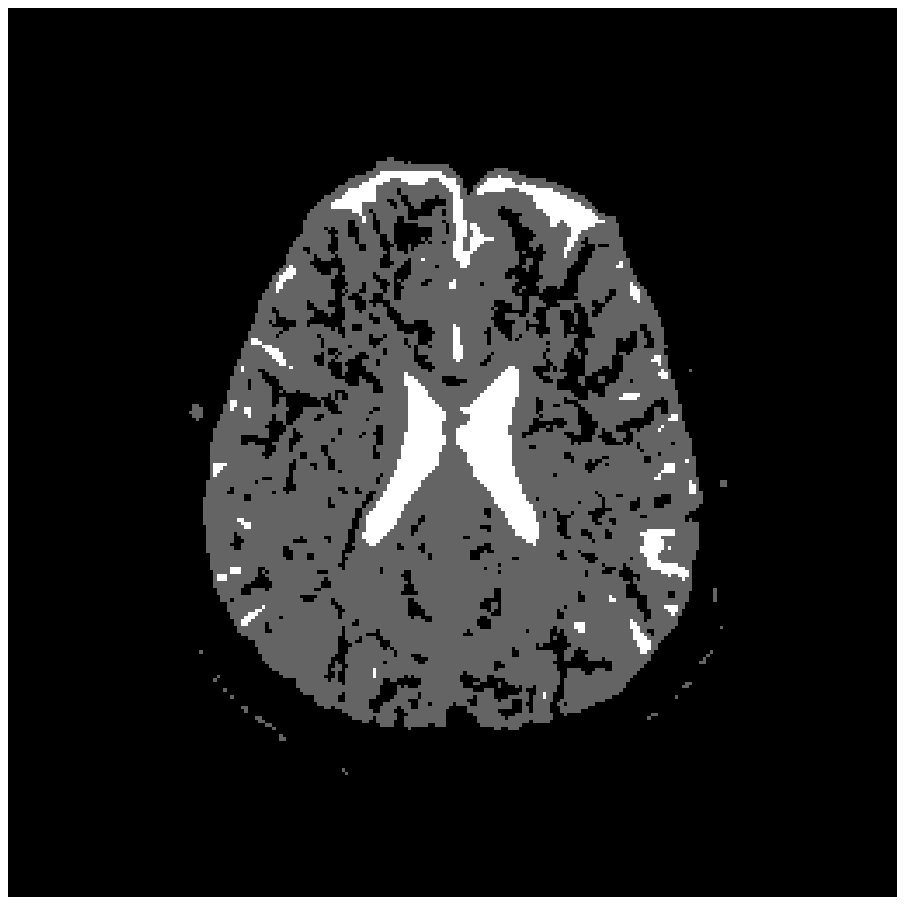}
	\caption{MLP classification of 13th slice with training set generated on ODC classification}
	\label{fig:classRNP4c_13}
\end{figure}

\begin{figure}
	\centering
		\includegraphics[width=0.45\textwidth]{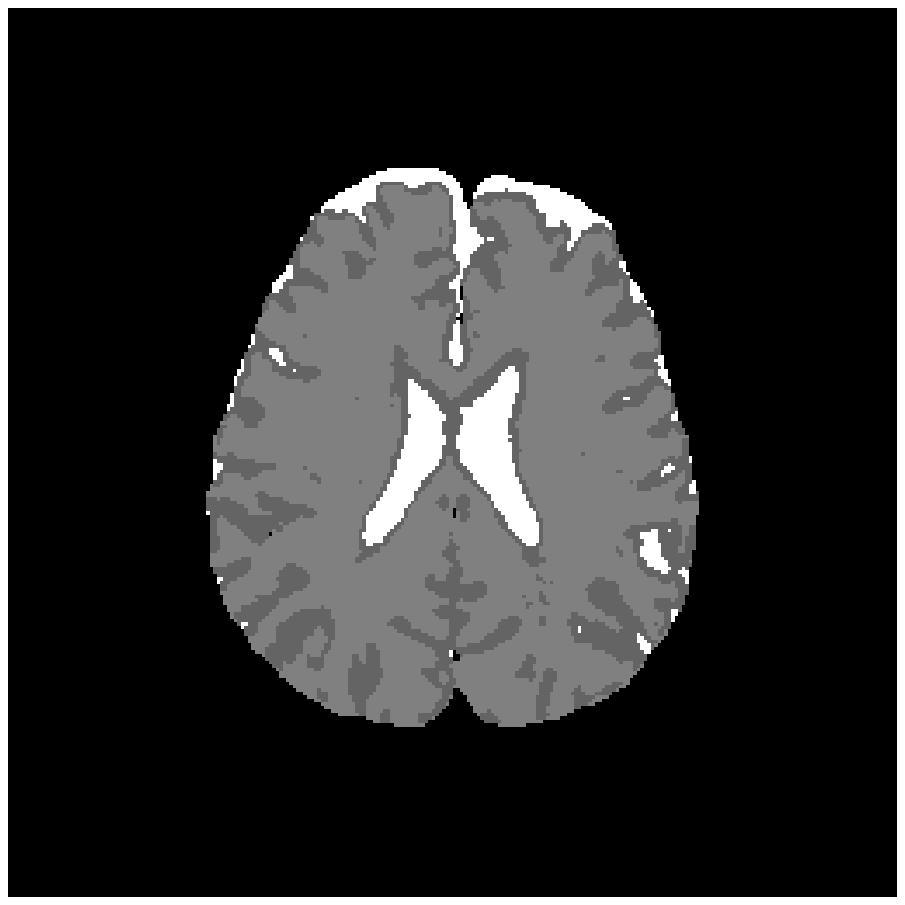}
	\caption{RBF classification of 13th slice with training set generated on ODC classification}
	\label{fig:classRBF4c_13}
\end{figure}

\begin{figure}
	\centering
		\includegraphics[width=0.45\textwidth]{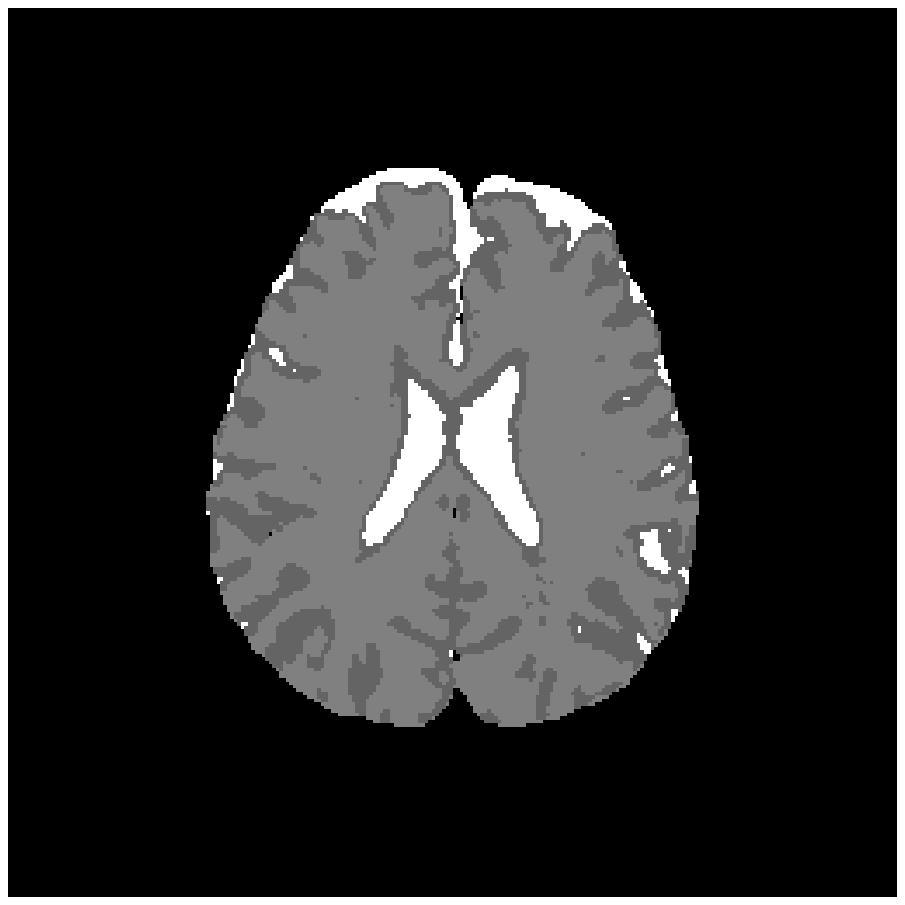}
	\caption{LVQ classification of 13th slice with training set generated on ODC classification}
	\label{fig:classLVQ4c_13}
\end{figure}

\begin{figure}
	\centering
		\includegraphics[width=0.45\textwidth]{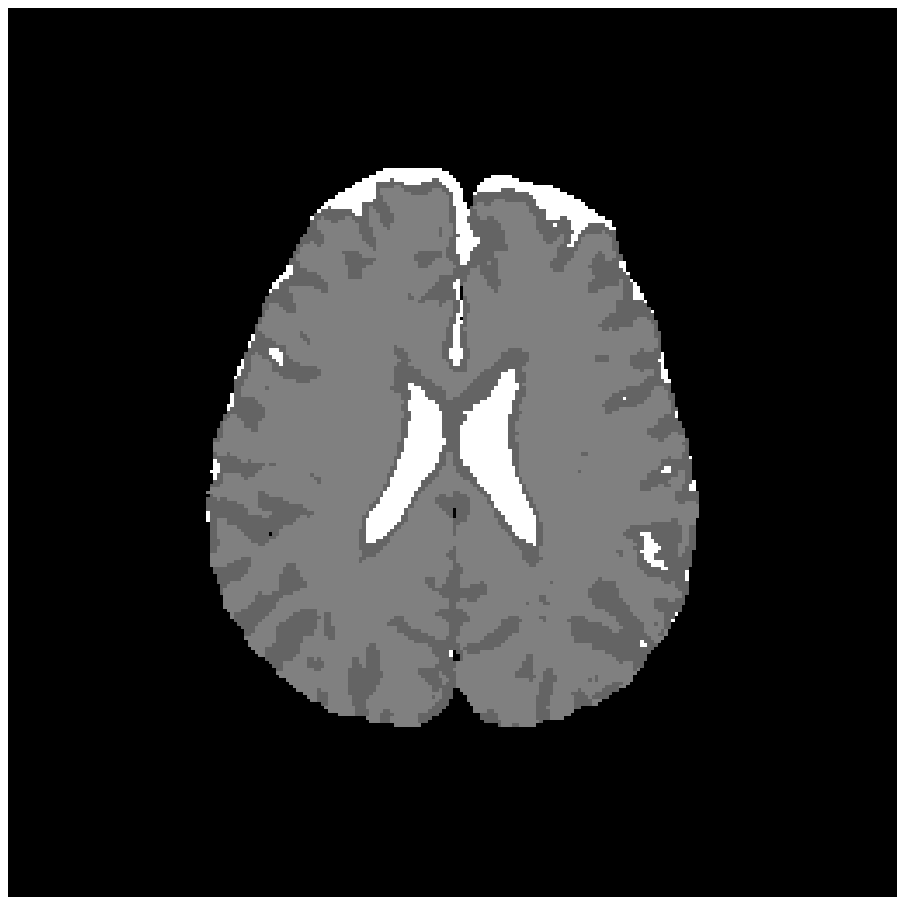}
	\caption{PO classification of 13th slice with training set generated on ODC classification}
	\label{fig:classPO4c_13}
\end{figure}

From the result obtained by ODC we can build a new training set (see figure \ref{fig:quadro_rois_treino4c}) and, consequently, training again the supervised classifiers. Table \ref{tab:ResultadosClassificacaoPosODC} shows the values of $\kappa$ and $\phi$ for the classification results of 13th slice by the supervised methods MLP, LVQ and RBF related to the result obtained by the use of method PO with 2-degree polynomials. Table \ref{tab:AreasClassificacaoPosODC} show the values of areas and fluid-matter ratio for 13th slice. Figures \ref{fig:classRNP4c_13}, \ref{fig:classRBF4c_13}, \ref{fig:classLVQ4c_13} and \ref{fig:classPO4c_13} show the results of the classification by methods MLP, RBF, LVQ and PO, respectively. It is evident that MLP network generated a classification where classes gray and white matter were classified as gray matter. However, the fluid-matter ratio is a bit close to the correct value, despite the low values of $\kappa$ and $\phi$ and the confusion between gray matter and image background. The fidelity of the results obtained by the use of methods MLP, RBF and LVQ to the ground truth image generated by PO classification was estimate using the morphological similarity index, resulting 0.1533, 0.9877 and 0.9877, respectively.

\begin{table}
    \caption{Overall accuracy $\phi$ (\%) and index $\kappa$ for the classification methods trained using ODC result}
  \begin{center}
  \begin{tabular} {cccc}
  \hline
  {} & {MLP} & {RBF} & {LVQ}\\
  \hline
  {$\phi$ (\%)} & {79.7699} & {98.6923} & {98.6923}\\
  {$\kappa$} & {0.5370} & {0.9701} & {0.9701}\\
  \hline
  \end{tabular}
  \end{center}
    \label{tab:ResultadosClassificacaoPosODC}
\end{table}

\begin{table}
    \caption{Percentual areas and fluid-matter ratio for the classification methods trained using ODC result}
  \begin{center}
  \begin{tabular} {ccccc}
  \hline
  {} & {MLP} & {RBF} & {LVQ} & {PO}\\
  \hline
  {$A_{\texttt{CSF}}$ (\%)} & {2.681} & {2.820} & {2.820} & {2.458}\\
  {$A_{\texttt{GM}}$ (\%)} & {21.603} & {6.496} & {6.496} & {6.682}\\
  {$A_{\texttt{WM}}$ (\%)} & {0.000} & {18.256} & {18.256} & {18.481}\\
  {$A_{\texttt{CSF}}/A_{\texttt{GM+WM}}$} & {0.124} & {0.114} & {0.114} & {0.098}\\
  \hline
  \end{tabular}
  \end{center}
    \label{tab:AreasClassificacaoPosODC}
\end{table}

\section{Discussion and Conclusions} \label{sec_discussion}

The results of the multispectral classification made possible a better quantitative and qualitative evaluation of acquired volumes, aiding at the measurement of the volumes of interest and the ratio between the volume occupied by cerebrospinal fluid and the volume of gray and white matter. Such a measurement can potentially aid to establish a correlation among this fluid-matter ratio, clinical variables and the evolution of damage caused by neurodegenerative diseases like Alzheimer's.

From table \ref{tab:ResultadosClassificacao} we can conclude that the multispectral approach, with index $\kappa$ of 0.6081, 0.9681, 0.9688, 0.9689 and 0.7755, for the multilayer perceptron, the radial basis function network, the Kohonen self-organized map, the Kohonen LVQ classifier, and the fuzzy c-means map, respectively, is superior to ADC map analysis by the use of the fuzzy c-means classifier, with index $\kappa$ of 0.2495.

These results are confirmed by the observation of the classified volumes of figures \ref{fig:VistaVolumeRNP}, \ref{fig:VistaVolumeRBF}, \ref{fig:VistaVolumeKO}, \ref{fig:VistaVolumeLVQ} and \ref{fig:VistaVolumeCM}, if compared to the classification of the ADC volume of figure \ref{fig:VistaVolumeCM_ADC}, showing several areas out of the region of the brain or inside the cranial box but marked as cerebrospinal fluid, white matter or gray matter.

From the obtained results it is evident that the multilayer perceptron and the fuzzy c-means classifier overestimated the volume occupied by cerebrospinal fluid (as seen in figures \ref{fig:VistaVolumeRNP} and \ref{fig:VistaVolumeCM}). When we compare such a result to the DW-MR volume with diffusion exponent 0 (see figure \ref{fig:VistaVolumet1}), left and right ventriculi are clearly separated. Furthermore, sulci were also overestimated. Such errors could contribute to influence specialists to evaluate neurodegenerative diseases as more advanced cases than they really are, as in present case of Alzheimer's disease.

The overestimation of volumes of interest is objectively perceived in results of table \ref{tab:AreasClassificacao}, where fluid-matter ratio ($V_1/V_2$) is 0.659 for the multilayer perceptron, and 0.450 for the fuzzy c-means classifier. These results are close to the result obtained by the classification of ADC maps by the fuzzy c-means classifier, with $V_1/V_2 = 0.546$. Hence these results are almost 6 times grater than fluid-matter ratio obtained from the radial basis function networks, with $V_1/V_2 = 0.094$, and from the Kohonen self-organized map and the Kohonen LVQ classifier, with $V_1/V_2 = 0.133$. These ratios are very close to the fluid-matter ratio obtained by the polynomial classification, $V_1/V_2 = 0,100$. Consequently, methods RBF, KO and LVQ can be considered good for the estimation of fluid-matter rate, whilst methods MLP and CM could be discharged as possible solutions to this nature of application.

The classification using the objective dialectical classifier demonstrated that the overestimation of the volumes of interest on classification results by the multilayer perceptron and the fuzzy c-means classifier has its origins in the association of classes liquor and gray matter. Obviously, such a situation is highly dependent upon the training set, mainly in the case of multilayer perceptron, once its classifications is supervised. Furthermore, multilayer perceptrons are considerably affected by local minima problems. Therefore, given a different training set, the resulting multilayer perceptron could associate gray and white matter to an unique class, similarly to the way the Kohonen LVQ classifier and the radial basis function network performed.

The results of the classification by multilayer perceptrons and fuzzy c-means classifiers, examples of supervised and non-supervised classifiers, respectively, showed that these classification approaches were not able to make distinction between gray matter and cerebrospinal fluid in several cases. This proves that there exist intermediate values of diffusion exponents between the values of gray matter's and cerebrospinal fluid's diffusion exponent. It proves that it is possible to differentiate such values. This is an interesting aspect, once it permits to extract more information from diffusion-weighted images than from the expected results provenient of ADC map analysis. This approach could optimize the extraction of information from diffusion-weighted images and reduce intrinsic costs, once new exams to extract anatomical information, like the acquisition of $T_1$, $T_2$ and PD-weighted images, can be unnecessary. New exams could be an upheaval for patients, even when submitted to non-invasive techniques like MRI. Furthermore, the longer the time patients are inside the tomographer, the greater the occurrence of errors, since it is perfectly possible for patients to make movements inside the tomographer, and sedating a patient is not easy and can have severe consequences.

Observing figure \ref{fig:gmDLT_13} indicating the detection of gray matter by the objective dialectical classifier, we can perceive the significative reduction of gray matter in brain frontal lobule, better seen in superior part of image. This lobule is directly related with the reduction of recent memory.

As a synthesis, we can conclude that the objective dialectical classifier could be a good mathematical tool to extract anatomical information from diffusion-weighted images, proving that it is possible to make distinction between gray and white matter, a difficult task to perform just using ADC maps.

The use of the objective dialectical classifier is good solution at performing classification tasks when it is needed to identify statistically significant classes and the initial number of classes is unknown. It makes possible the detection of relevant classes and singularities not initially predicted. This could aid the specialist to measure volumes of interest and to try to establish a correlation between determined anatomical aspects and the progress of neurodegenerative diseases like Alzheimer's, for instance. Such an evolvable classification could also aid the specialist in cases where it is specially necessary to detect significant alterations of diffusion coefficient measured values, for instance, potentially improving both quantitative and qualitative the analysis of DW-MR images.

\bibliographystyle{unsrt}
\bibliography{arq_bib}

\begin{thebibliography}{10}

\bibitem{marx1980_1}
K.~{Marx}.
\newblock Critique of {Hegel}'s dialectics and philosophy.
\newblock In {\em Economic and Philosophic Manuscripts of 1844}. International
  Publishers, 1980.

\bibitem{engels1975}
F.~{Engels}.
\newblock The role played by labor in the transition from ape to man.
\newblock In {\em Collected Works of Karl Marx and Frederik Engels}.
  International Publishers, 1975.

\bibitem{gramsci1992_1}
A.~{Gramsci}.
\newblock {Introduction to the Study of Philosophy and Historical Materialism}.
\newblock In {\em Prison Notebooks}. Columbia University, 1992.

\bibitem{gramsci1992_2}
A.~{Gramsci}.
\newblock {Some Problems in the Study of the Philosophy of Praxis}.
\newblock In {\em Prison Notebooks}. Columbia University, 1992.

\bibitem{bobbio1990}
N.~{Bobbio}.
\newblock {\em Saggi su Gramsci}.
\newblock Feltrinelli, Milano, 1990.

\bibitem{thornley2007}
C.~{Thornley} and F.~{Gibb}.
\newblock {A dialectical approach to information retrieval}.
\newblock {\em Journal of Documentation}, 63(5):755--764, 2007.

\bibitem{rosser2000}
J.~B. {Rosser Jr.}
\newblock {Aspects of dialectics and nonlinear dynamics}.
\newblock {\em Cambridge Journal of Economics}, 24(3):311--324, 2000.

\bibitem{ewers2006}
M.~{Ewers}, S.~J. {Teipel}, O.~{Dietrich}, S.~O. {Schönberg}, F.~{Jessen},
  R.~{Heun}, P.~{Scheltens}, L.~{van de Pol}, N.~R. {Freymann}, H.~J.
  {Moeller}, and H.~{Hampela}.
\newblock {Multicenter assessment of reliability of cranial MRI}.
\newblock {\em Neurobiology of Aging}, (27):1051--1059, 2006.

\bibitem{carmichael2005}
O.~T. {Carmichael}, H.~A. {Aizenstein}, S.~W. {Davis}, J.~T. {Becker}, P.~M.
  {Thompson}, C.~C. {Meltzer}, and Y.~{Liu}.
\newblock {Atlas-based hippocampus segmentation in Alzheimer's disease and mild
  cognitive impairment}.
\newblock {\em NeuroImage}, (27):979--990, 2005.

\bibitem{hirata2005}
Y.~{Hirata}, H.~{Matsuda}, K.~{Nemoto}, T.~{Ohnishi}, K.~{Hirao},
  F.~{Yamashita}, T.~{Asada}, S.~{Iwabuchi}, and H.~{Samejima}.
\newblock {Voxel-based morphometry to discriminate early Alzheimer's disease
  from controls}.
\newblock {\em Neuroscience Letters}, (382):269--274, 2005.

\bibitem{pannacciulli2006}
N.~{Pannacciulli}, A.~{Del Parigi}, K.~{Chen}, D.~S. N.~T. {Le}, E.~M.
  {Reiman}, and P.~A. {Tataranni}.
\newblock Brain abnormalities in human obesity: A voxel-based morphometric
  study.
\newblock {\em NeuroImage}, (31):1419--1425, 2006.

\bibitem{friman2006}
O.~{Friman}, G.~{Farnebäck}, and C.~F. {Westin}.
\newblock A bayesian approach for stochastic white matter tractography.
\newblock {\em IEEE Transactions on Medical Imaging}, 25(8):965--978, 2006.

\bibitem{naggara2006}
O.~{Naggara}, C.~{Oppenheim}, D.~{Rieu}, N.~{Raoux}, S.~{Rodrigo}, G.~D.
  {Barba}, and J.~F. {Meder}.
\newblock {Diffusion tensor imaging in early Alzheimer's disease}.
\newblock {\em Psychiatry Research Neuroimaging}, (146):243--249, 2006.

\bibitem{bozzali2002}
M.~{Bozzali}, A.~{Falini}, M.~{Franceschi}, M.~{Cercignani}, M.~{Zuffi},
  G.~{Scotti}, G.~{Comi}, and M.~{Filippi}.
\newblock White matter damage in alzheimer's disease assessed in vivo using
  diffusion tensor magnetic resonance imaging.
\newblock {\em Journal of Neurology, Neurosurgery and Psychiatry}, 72:742--746,
  2002.

\bibitem{du2005}
A.~T. {Du}, N.~{Schuff}, L.~L. {Chao}, J.~{Kornak}, F.~{Ezekiel}, W.~J.
  {Jagust}, J.~H. {Kramer}, B.~R. {Reed}, B.~L. {Miller}, D.~{Norman}, H.~C.
  {Chui}, and M.~W. {Weiner}.
\newblock White matter lesions are associated with cortical atrophy more than
  entorhinal and hippocampal atrophy.
\newblock {\em Neurobiology of Aging}, (26):553--559, 2005.

\bibitem{haacke1999}
E.~M. {Haacke}, R.~W. {Brown}, M.~R. {Thompson}, and R.~{Venkatesan}.
\newblock {\em Magnetic Resonance Imaging: Physical Principles and Sequence
  Design}.
\newblock Wiley-Liss, 1999.

\bibitem{hayasaka2006}
S.~{Hayasaka}, A.~T. {Du}, A.~{Duarte}, J.~{Kornak}, G.~H. {Jahng}, M.~W.
  {Weiner}, and N.~Schuff.
\newblock {A non-parametric approach for co-analysis of multi-modal brain
  imaging data: Application to Alzheimer's disease}.
\newblock {\em NeuroImage}, (30):768--779, 2006.

\bibitem{santos2007d}
W.~P. {Santos}, R.~E. {Souza}, and P.~B. {Santos Filho}.
\newblock {Evaluation of Alzheimer's disease by analysis of MR images using
  multilayer perceptrons and Kohonen SOM classifiers as an alternative to the
  ADC maps}.
\newblock In {\em 29th Annual International Conference of the IEEE Engineering
  in Medicine and Biology Society}, Lyon, France, 2007. EMBS-IEEE.

\bibitem{santos2008r}
W.~P. {Santos}, R.~E. {Souza}, A.~F.~D. {Silva}, and P.~B. {Santos Filho}.
\newblock {Evaluation of Alzheimer's disease by analysis of MR images using
  multilayer perceptrons and committee machines}.
\newblock {\em Computerized Medical Imaging and Graphics}, 32(1):17--21, 2008.

\bibitem{santos2007a}
W.~P. {Santos}, R.~E. {Souza}, A.~F.~D. {Silva}, and P.~B. {Santos Filho}.
\newblock {Evaluation of Alzheimer's disease by analysis of MR images using
  multilayer perceptrons, polynomial nets and Kohonen LVQ classifiers}.
\newblock In {\em Lecture Notes in Computer Science: Computer Vision / Computer
  Graphics Collaboration Techniques and Applications (MIRAGE 2007)}, volume~2,
  pages 12--22, Rocquencourt, France, 2007. INRIA \& CS-IEEE.

\bibitem{santos2007b}
W.~P. {Santos}, R.~E. {Souza}, A.~F.~D. {Silva}, and P.~B. {Santos Filho}.
\newblock {Avaliação da Doença de Alzheimer pela Análise Multiespectral de
  Ima\-gens DW-MR por Mapas Auto-Organizados de Kohonen como Alternativa aos
  Mapas ADC}.
\newblock In {\em IV Congreso Latino-Americano de Engenharia Biomédica},
  Porlamar, Venezuela, 2007. IFMBE.

\bibitem{santos2007c}
W.~P. {Santos}, R.~E. {Souza}, A.~F.~D. {Silva}, and P.~B. {Santos Filho}.
\newblock {Avaliação da doença de Alzheimer pela análise multiespectral de
  imagens DW-MR por redes RBF como alternativa aos mapas ADC}.
\newblock In {\em VIII Congresso Brasileiro de Redes Neurais}, Florianópolis,
  Brasil, 2007. SBRN.

\bibitem{santos2006a}
W.~P. {Santos}, R.~E. {Souza}, A.~F.~D. {Silva}, N.~M. {Portela}, and P.~B.
  {Santos Filho}.
\newblock {Análise multiespectral de imagens cerebrais de ressonância magnética
  ponderadas em difusão usando lógica nebulosa e redes neurais para avaliação
  de danos causados pela doença de Alzheimer}.
\newblock In {\em XI Congresso Brasileiro de Física Médica}, Ribeirão Preto,
  Brasil, 2006. Sociedade Brasileira de Física Médica.

\bibitem{santos2006b}
W.~P. {Santos}, R.~E. {Souza}, A.~F.~D. {Silva}, N.~M. {Portela}, and P.~B.
  {Santos Filho}.
\newblock {Avaliação da doença de Alzheimer por análise de imagens de RMN
  utilizando redes MLP e máquinas de comitê}.
\newblock In {\em XX Congresso Brasileiro de Engenharia Biomédica}, São Pedro,
  Brasil, 2006. Sociedade Brasileira de Engenharia Biomédica.

\bibitem{wu2001}
W.~C. {Wu}, K.~H. {Kao}, P.~H. {Lai}, and H.~W. {Chung}.
\newblock {Multi-Component Decay Behavior on High-b-Value Diffusion Weighted
  MRI}.
\newblock In {\em Procedings of the 23rd Annual EMBS International Conference},
  pages 2286--2288, Instanbul, Turkey, 2001. EMBS-IEEE.

\bibitem{elshafiey2002}
I.~{Elshafiey}.
\newblock Diffusion tensor magnetic magnetic resonance imaging of lesions in
  multiple sclerosis patients.
\newblock In {\em 19th National Radio Science Conference}, pages 626--633,
  Alexandria, Egypt, 2002. EMBS-IEEE.

\bibitem{basser2003}
P.~J. {Basser} and S.~{Pajevic}.
\newblock {A Normal Distribution for Tensor-Valued Random Variables:
  Applications to Diffusion Tensor MRI}.
\newblock {\em IEEE Transactions on Medical Imaging}, 22(7):785--794, 2003.

\bibitem{wang2004}
B.~{Wang}, P.~K. {Saha}, J.~K. {Udupa}, M.~A. {Ferrante}, J.~{Baumgardner},
  D.~A. {Roberts}, and R.~R. {Rizi}.
\newblock {3D airway segmentation method via hyperpolarized $^3$He gas MRI by
  using scale-based fuzzy connectedness}.
\newblock {\em Computerized Medical Imaging and Graphics}, (28):77--86, 2004.

\bibitem{chen2004}
Y.~{Chen}, W.~{Guo}, Q.~{Zeng}, X.~{Yan}, F.~{Huang}, H.~{Zhang}, G.~{He},
  B.~C. {Vemuri}, and Y.~{Liu}.
\newblock {Estimation, Smoothing, and Characterization of Apparent Diffusion
  Coefficient Profiles from High Angular Resolution DWI}.
\newblock In {\em Proceedings of the 2004 IEEE Computer Society Conference on
  Computer Vision and Pattern Recognition}. CS-IEEE, 2004.

\bibitem{guo2006}
W.~{Guo}, Q.~{Zeng}, Y.~{Chen}, and Y.~{Liu}.
\newblock Using multiple tensor detection to reconsctruct white matter fiber
  traces with branching.
\newblock In {\em Proceedings of the ISBI 2006}. CS-IEEE, 2006.

\bibitem{maraga2006}
C.A. {Castano-Moraga}, C.~{Lenglet}, R.~{Deriche}, and J.~{Ruiz-Alzola}.
\newblock {A Fast and Rigorous Anisotropic Smoothing Method for DT-MRI}.
\newblock In {\em Proceedings of the ISBI 2006}. CS-IEEE, 2006.

\bibitem{liang2000}
Z.~P. {Liang} and P.~C. {Lauterbur}.
\newblock {\em Principles of Magnetic Resonance Imaging: A Signal Processing
  Perspective}.
\newblock IEEE Press, New York, 2000.

\bibitem{gullberg1999}
G.~T. {Gullberg}, D.~G. {Roy}, G.~L. {Zeng}, A.~L. {Alexander}, and D.~L.
  {Parker}.
\newblock {Tensor Tomography}.
\newblock {\em IEEE Transactions on Nuclear Science}, 46(4):991--1000, 1999.

\bibitem{basser2002}
P.~J. {Basser}.
\newblock {Diffusion-Tensor MRI: Theory, Experimental Design, and Data
  Analysis}.
\newblock In {\em Procedings of the 2nd Joint EMBS BMES Conference}, pages
  1165--1166, Houston, USA, 2002. EMBS-IEEE-BMES.

\bibitem{kang2005}
N.~{Kang}, J.~{Zhang}, E.~S. {Carlson}, and D.~{Gembris}.
\newblock White matter fiber tractography via anisotropic diffusion simulation
  in the human brain.
\newblock {\em IEEE Transactions on Medical Imaging}, 24(9):1127--1137, 2005.

\bibitem{fillard2006}
P.~{Fillard}, V.~{Arsigny}, X.~{Pennec}, and N.~{Ayache}.
\newblock {Clinical DT-MRI estimations, smoothing and fiber tracking with
  log-Euclidean metrics}.
\newblock In {\em Proceedings of the ISBI 2006}. CS-IEEE, 2006.

\bibitem{santos2008a}
W.~P. {Santos}, R.~E. {Souza}, P.~B. {Santos Filho}, F.~B. {Lima Neto}, and
  F.~M. {Assis}.
\newblock {A Dialectical Approach for Classification of DW-MR Alzheimer's
  Images}.
\newblock In {\em IEEE World Congress on Computational Intelligence (WCCI
  2008)}, Hong Kong, China, 2008. CIS-IEEE.

\bibitem{santos2008b}
W.~P. {Santos}, F.~M. {Assis}, R.~E. {Souza}, and P.~B. {Santos Filho}.
\newblock {Evaluation of Alzheimer's Disease by Analysis of MR Images using
  Objective Dialectical Classifiers as an Alternative to ADC Maps}.
\newblock In {\em 30th Annual International Conference of the IEEE Engineering
  in Medicine and Biology Society}, Vancouver, Canada, 2008. EMBS-IEEE.

\bibitem{duda2001}
R.~{Duda}, P.~{Hart}, and D.~G. {Stork}.
\newblock {\em Pattern Classification}.
\newblock John Wiley and Sons, 2001.

\bibitem{duda1972}
R.~{Duda} and P.~{Hart}.
\newblock {\em Pattern Classification and Scene Analysis}.
\newblock John Wiley and Sons, 1972.

\bibitem{sklansky1981}
J.~{Sklansky} and G.~N. {Wassel}.
\newblock {\em Pattern Classifiers and Trainable Machines}.
\newblock Springer-Verlag, 1st edition, 1981.

\bibitem{haykin1999}
S.~{Haykin}.
\newblock {\em Neural Networks: A Comprehensive Foundation}.
\newblock Prentice Hall, New York, 1999.

\bibitem{ulisses1994}
U.~M. {Braga Neto}.
\newblock Reconstrução volumétrica e análise de imagens tridimensionais por
  morfologia matemática.
\newblock Master's thesis, Faculdade de Engenharia Elétrica da Universidade
  Estadual de Campinas, Campinas, Brasil, 1994.

\bibitem{landgrebe2002}
D.~{Landgrebe}.
\newblock Hyperspectral image analysis.
\newblock {\em IEEE Signal Processing Magazine}, Jan 2002.

\bibitem{wang2002}
Z.~{Wang} and A.~C. {Bovik}.
\newblock A universal image quality index.
\newblock {\em IEEE Signal Processing Letters}, 9, 2002.

\end{thebibliography}









\end{document}